\pdfoutput=1

\documentclass[11pt]{article}

\usepackage{acl}

\usepackage{algorithm}
\usepackage{algorithmicx}
\usepackage{algpseudocode}
\usepackage{times}
\usepackage{latexsym}
\usepackage{amsmath}
\usepackage{graphicx}
\usepackage{color}
\usepackage{colortbl}
\usepackage{booktabs}
\usepackage{url}
\usepackage{multirow}
\usepackage[T1]{fontenc}
\usepackage[utf8]{inputenc}
\usepackage{microtype}
\usepackage{CJKutf8}
\usepackage{tabularx}
\usepackage{graphicx} 
\usepackage{hyperref}
\usepackage{amssymb}

\title{PlanGPT: Enhancing Urban Planning with Tailored Language Model and Efficient Retrieval}

 \author{
  $^1$ He Zhu , $^1$ Wenjia Zhang \footnote{Corresponding author , wenjiazhang@pku.edu.cn} , $^1$Nuoxian Huang , $^1$Boyang Li , $^1$Luyao Niu , $^4$Zipei Fan \\ 
  \textbf{$^1$Tianle Lun , $^1$Yicheng Tao , $^1$Junyou Su , $^1$Zhaoya Gong , $^{23}$Chenyu Fang , $^2$Xing Liu }  \\
  $^1$ Behavioral and Spatial AI Lab , Peking University \\
  $^2$ China Academy of Urban Planning \& Design \\
  $^3$ Technical University of Munich \and $^4$University of Tokyo   \\
  zhuye140@gmail.com , wenjiazhang@pku.edu.cn
 }

\begin{document}
\maketitle

\renewcommand{\thefootnote}{\fnsymbol{footnote}}
\footnotetext[1]{Corresponding author , wenjiazhang@pku.edu.cn}
\renewcommand{\thefootnote}{\arabic{footnote}}

\begin{abstract}
In the field of urban planning, general-purpose large language models often struggle to meet the specific needs of planners. Tasks like generating urban planning texts, retrieving related information, and evaluating planning documents pose unique challenges. To enhance the efficiency of urban professionals and overcome these obstacles, we introduce \textbf{PlanGPT}, the first specialized Large Language Model tailored for urban and spatial planning. Developed through collaborative efforts with institutions like the Chinese Academy of Urban Planning, PlanGPT leverages a customized local database retrieval framework, domain-specific fine-tuning of base models, and advanced tooling capabilities. Empirical tests demonstrate that PlanGPT has achieved advanced performance, delivering responses of superior quality precisely tailored to the intricacies of urban planning.
\end{abstract}

\section{Introduction} \label{introduction}

Due to the impressive reasoning, memory, and comprehension abilities inherent in large language models\cite{chatgpt}, substantial progress and prospects have arisen in various domains. Particularly in fields like finance, medicine, and law, customized large models tailored to specific verticals have emerged, efficiently tackling challenges issues commonly associated with general-purpose large models, such as vague responses and hallucinations caused by uniform training data distribution, thereby boosting staff productivity.

\begin{figure}[t]
  \centering
  \includegraphics[width=0.5\textwidth]{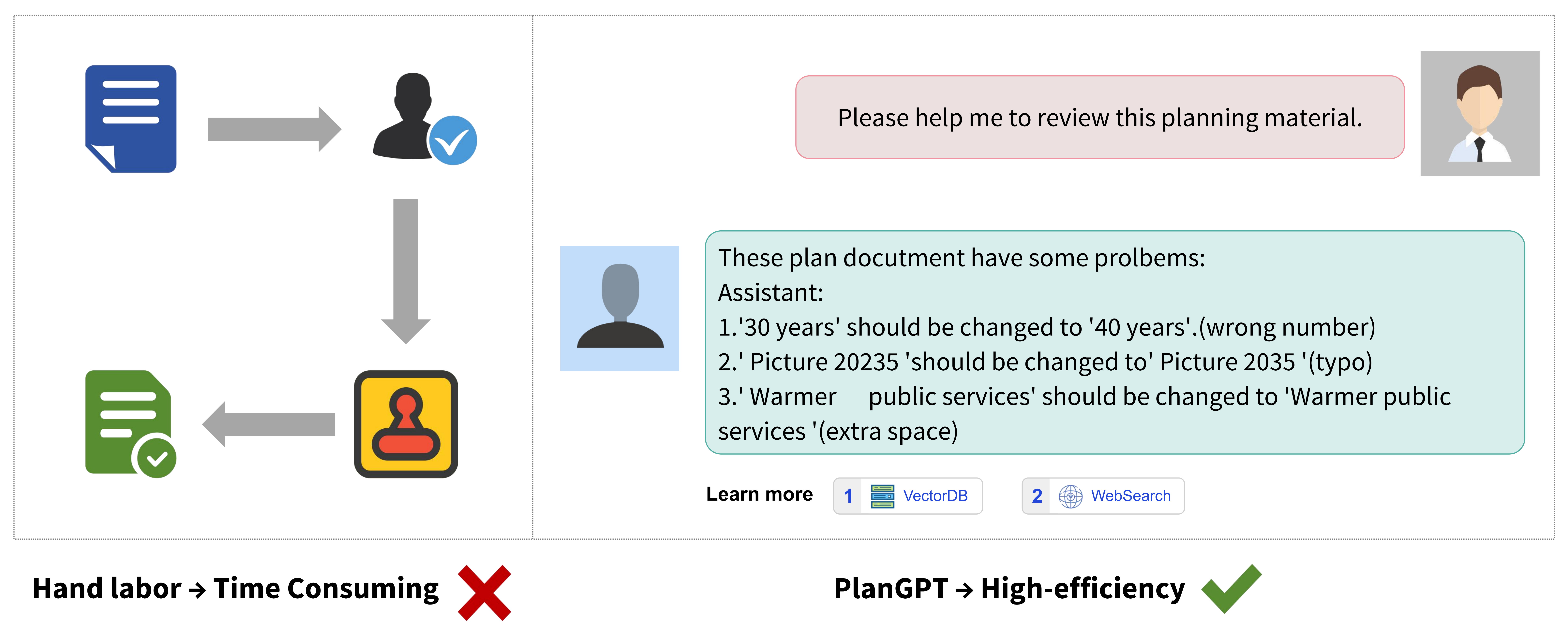}
  \caption{Review task workflow}
  \label{usage1}
\end{figure}

Through discussions with planners from city planning departments/companies, it became evident that significant amounts of time are expended on tasks such as planning text management, review, audit, and assessment. For instance, during text review, staff meticulously evaluate each item against a standard framework, rectifying errors or omissions in urban planning documents. Similarly, in text assessment, staff evaluate documents from multiple dimensions (legality, feasibility, economic viability, innovativeness), which consume considerable time and effort. Leveraging the robust comprehension and reasoning abilities of LLMs, we posit that the aforementioned processes can be addressed through the incorporation of large language model, as shown in Figure \ref{usage1}.

However, in practical operations, we have found that it is not an easy task due to the inherent nature of the Chinese urban planning industry and the characteristics of urban planning texts:

\begin{enumerate}
    \item \textbf{Government document style:} Linked to government affairs, urban planning documents often employ fixed phrases and structures, creating a challenge for LLMs to balance government style with informative content. The low signal-to-noise ratio in these documents complicates retrieval. Moreover, heightened attention to data security restricts model selection \footnote{The models are suggested to open-source and \textbf{CSPON}-compliant. (China Spatial Planning Online Monitoring Network).}.

    \item \textbf{Interdisciplinary knowledge:} Urban and spatial planning texts integrate knowledge from multiple disciplines such as environmental science, ecology, economics, and law. However, current large models have not effectively activated knowledge in this specialized field, making it difficult to utilize them effectively.

    \item \textbf{Timeliness and multimodality:} Urban planning documents require synchronization with government regulations and are laden with images and tabular data, necessitating specialized tools for analysis and processing.
\end{enumerate}

To address the distinctive challenges inherent in urban planning texts, we present the first Large Language Model in the urban planning domain: \textbf{PlanGPT}. Firstly, it features a customized embedding model and vector database retrieval system for accurate information extraction in vast amounts of urban planning texts, overcoming the low signal-to-noise ratio characteristic of the urban planning domain by using keyword extraction and hierarchical search techniques. Additionally, we employ instruction fine-tuning methods to activate the model's interdisciplinary knowledge and enhance its proficiency in mastering the style of governmental documents, meeting the demands of planners.  Furthermore, inspired by advancements in agent-based systems within the realm of large models, PlanAgent has been created to strategically utilize resources like networks, visual aids, charts, or domain-specific models. This approach significantly tackles the issues related to timeliness and multimodality in planning documents.

Experimental results have demonstrated that PlanGPT effectively addresses all the aforementioned challenges, fulfilling the needs of planners in the four typical tasks of daily work, surpassing other state-of-the-art models.

\begin{figure*}[t]
      \centering
      \includegraphics[width=\textwidth]{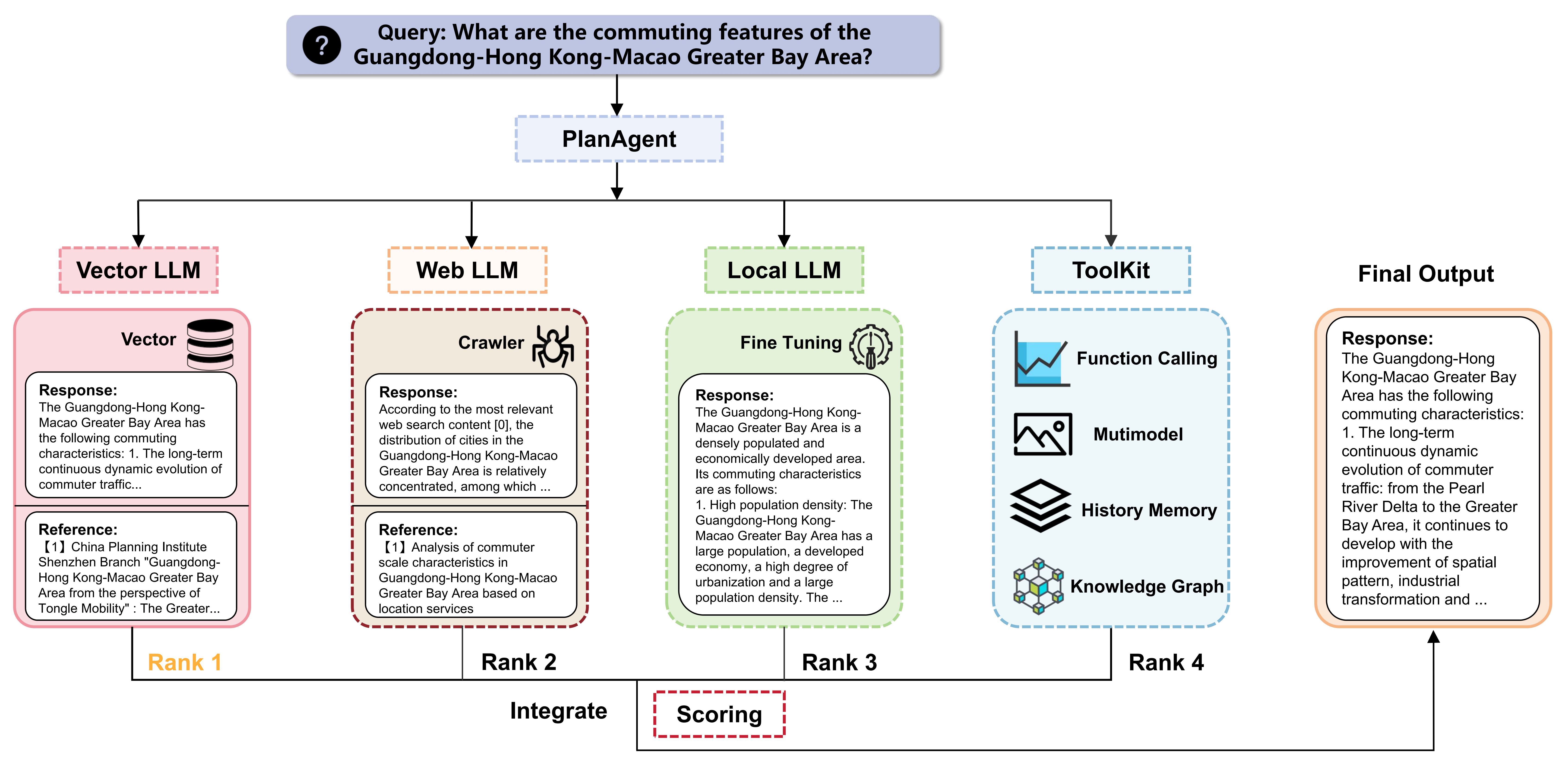}
      \caption{PlanGPT Architecture}
      \label{plangpt}
\end{figure*}

\section{Related Works}

\subsection{General-Purpose and Vertical-Specific Large Language Models}
Large language models (LLMs) encompass both general-purpose and vertical-specific applications, showcasing their versatility and effectiveness. Notable models like ChatGPT\cite{chatgpt}, GPT-4\cite{GPT4}, LLaMA\cite{touvron2023llama} series\cite{touvron2023llama}, Bard\cite{bard}, PaLM2\cite{PaLM2}, Claude2\cite{Claude2}, Mistral\cite{mistral} and Gemini\cite{gemini}, demonstrate broad capabilities across various tasks and industries. In the Chinese language domain, models like the Baichuan series\href{https://www.baichuan-ai.com/home}, GLM series\cite{GLM}, Kimi-chat\href{https://kimi.moonshot.cn/}, Yi\href{https://github.com/01-ai/Yi}, Qwen\href{https://tongyi.aliyun.com}, Skywork\cite{wei2023skywork} and LLaMA-Chinese\cite{Chinese-LLaMA-Alpaca} offer several advantages tailored to the Chinese language and its unique challenges.
Vertical-specific applications also benefit from LLMs. Examples include HuaTuo\cite{huatuo}, a medical domain model, and ChatLaw\cite{ChatLaw}, an open-source legal LLM, which address specific needs within their respective domains. Similarly, XuanYuan 2.0\cite{XuanYuan} caters to the finance sector, DoctorGLM\cite{Doctorglm} focuses on healthcare, and MathGPT\cite{MathGPT} enhances mathematical problem-solving capabilities. These models collectively highlight the diverse applications and potential of LLMs across different domains.

\subsection{Language Models in urban planning domain}
In the fields relevant to urban planning such as geography and transportation, several specialized models have emerged. TrafficGPT\cite{TrafficGPT} integrates ChatGPT with traffic foundation models to enhance urban traffic management and decision support through data analysis and natural language dialogues. Prithvi\cite{Prithvi-100M}, a NASA-derived model, focuses on climate, disaster, and geography predictions, pre-trained on IBM's watsonx.ai, serving applications like climate change, flood mapping, and crop yield forecasting. TransGPT\cite{TransGPT}, as China's first open-source traffic model, finds applications in traffic prediction, advisory, public transport services, urban planning, safety education, accident analysis, and autonomous driving support. EarthGPT\cite{earthgpt}, a multi-modal large language model (MLLM) designed for remote sensing (RS) images, integrates RS interpretation tasks to enhance both visual perception and language understanding. Currently, there is no large model specifically tailored for urban and spatial planning domain, so we humbly introduce PlanGPT to address this gap.

\subsection{Mitigation of hallucination}
In vertical domains, the faithfulness and factualness of large model outputs are heavily reliant. Retrieval techniques, fine-tuning methods, and agent tools have been proven to effectively mitigate model hallucination issues. RAG combines parameterized knowledge from LLMs with non-parameterized external knowledge to alleviate hallucination problems. Outstanding retrieval works such as Raven\cite{huang2023raven}, Retro\cite{borgeaud2022improving}, Toc Sugre\cite{kim2023tree}, selfmem\cite{cheng2024lift}, genread\cite{yu2022generate}, and RECITE\cite{sun2022recitation} contribute significantly in this regard.  Notably, Self-RAG\cite{asai2023self} framework introduces a retrieval token to determine whether to recall documents, followed by assessing document validity using a critique token. FLARE\cite{jiang2023active} iteratively executes retrieval, judging the need for answer regeneration based on probability calculations. RA-DIT\cite{lin2023ra} enhances LM's use of retrieved information and refines the retriever for more relevant results, yielding significant performance gains when combined.
Instruction fine-tuning significantly enhances model capabilities and effectively alleviates hallucinations. By employing methods like humpback\cite{li2023self}, kun\cite{zheng2024kun}, and muffin\cite{lou2023muffin}, we collect data from various sources, ensuring quality through filtering methods like deita\cite{liu2024what}, cherry\cite{Li2023FromQT}, mods\cite{du2023mods}, etc. Additionally, techniques such as wizardlm\cite{xu2023wizardlm} and self-instruct\cite{wang2022self} increase data difficulty, improving model robustness. Agents can determine the appropriate tools to use, such as web searches(webglm\cite{webglm},webgpt\cite{nakano2021webgpt}) or function calls, to enhance the quality of model outputs. Inspired by these work, we have innovatively proposed retrieval and instruction labeling methods tailored for urban planning domains, in conjunction with PlanAgent, effectively mitigating hallucination issues in large models.

\section{PlanGPT}

In this section, we will introduce the overarching framework and technical intricacies of PlanGPT.

\subsection{Vector-LLM}

    \par In urban planning, professionals often struggle to find relevant materials from large datasets. This task can be modeled as the identification of the most pertinent document span \( s^* \) within a collection \( s \), defined as \( s^* = \arg\max_{s \in S} \text{Relate}(q, s) \), where \( \text{Relate}(q, s) \) represents the similarity function between inquiry \( q \) and document span \( s \).

    \subsubsection{{\textbf{Plan-Emb}}} \label{planemb-section}
    Advanced embedding methods are considered common solutions in enhancing semantic understanding, but they still produce suboptimal results in the field of urban planning due to two reasons: \textbf{(1) Specialized Terminology}: Urban planning possesses its own linguistic system, characterized by abbreviations and substitutions for specialized terms. For example, \textit {regulations} may refer to \textit{zoning regulations}, \textit{land type} to \textit{land use classification}, causing ambiguity, especially in Chinese. \textbf{(2) Planner's Perspective on Vocabulary}: Common terms like \textit{land use} carry richer meanings for planners. While commonly understood as land utilization, planners view it as interactions between people, land, and ecosystems. This difference in perspective affects semantic understanding and search accuracy.

    Drawing inspiration from previous work involving embedding models\cite{cui2022lert, cui-etal-2021-pretrain, mikolov2013efficient, chen2024bge, reimers2019sentencebert, gao2022simcse,kexuefm-8847}, we introduce our embedding model \textbf{Plan-Emb} for urban planning domain. \textbf{Plan-Emb} is an embedding model tailored for comprehending urban-planning-specific knowledge with two-stage training process: initial pre-training using general Chinese text labels\cite{snli:emnlp2015} , followed by supervised fine-tuning on self-collected urban planning datasets. A regularization InfoNCE loss\cite{oord2018representation} is introduced during the second stage to prevent catastrophic forgetting of prior model capabilities. 
    \[
    \text{loss} = - \log \frac{e^{\text{sim}(h^{q}, h^{a^{+}}) / \tau }}{\sum_{i=0}^{N} e^{\text{sim}(h^{q}, h^{a_i}) / \tau} } + \lambda D_{KL}(P || Q)
    \]
    where \(q\) and \(a^{+}\) represent the sentence and its positive samples, while \(P\) and \(Q\) denote the model's output distributions after pre-train and fine-tuning stage, respectively.

    For fine-tuning data collection, we initially leverage LLMs to filter keywords or key sentences aligned with our self-curated teaching syllabus. Subsequently, a cost-effective approach involving perturbations, explanations, and rewriting is employed to generate positive samples. Following experiments have confirmed the effectiveness of PlanEmb.

    \subsubsection{{\textbf{Plan-HS}(Hierarchical Search)}} \label{planhs-section}
    
    To address the challenges of low signal-to-noise ratio and declining embedding capability with longer sentences, we introduce a novel hierarchical embedding approach for query processing (depicted in Algorithm \ref{hs_search}). 
    In the data pre-processing phase, tailored keywords extraction method (PlanKeyBert) is employed to extract relevant keywords \(d_i\) from input document \(D\) and store them in a hash-map, mapping each chunk \(d_i\) to its corresponding \(k_i\) while retaining essential information. During the search process, a query \(Q\) is used to recall relevant documents from vectorDB based on keyword and semantic similarity scores. Subsequently, hard matching scores and advanced cross-attention scores are employed to rerank the recall results.

 \subsection{Local-LLM} \label{localllm}
    \par Large language models often struggle to integrate domain-specific knowledge, such as in urban planning, leading to language generation that deviates from established conventions.
    The challenge here lies not solely in the absence of domain-specific data\footnote{ It has been observed that a significant portion of pre-trained data in general large-scale models already encompasses domain-specific data related to urban planning. }, but rather in the model's incapacity to synthesize and apply knowledge within this specialized domain.
    
    To address these challenges, we conducted a two-stage model adaptation: Urban planning Knowledge Activation and Specific Capability Development.

      \begin{algorithm}
    \caption{Hierarchical Search}\label{hs_search}
    \begin{algorithmic}[1]

    \Procedure{PreProcess}{}
        \State Initialize the PlanKeyBERT and PlanEmb
        \State Initialize the vector database $V$ for chunks and hash-mapper $H$ to store the chunks and keywords, where:
        \begin{align*}
        D &= \{d_1, d_2, \ldots, d_n\} \\
        V &: D \rightarrow \mathbb{R}^m \\
        H &: \{d_i\} \rightarrow \{K_{i}\}
        \end{align*}
    \EndProcedure
    
    \Procedure{QuerySearch}{query}
        \State Obtain embedding vector $s$ and keywords $K$ for the query.
        \State Recall Top($x/2$) chunks by $sim(K, K_i)$ assign to \textbf{A}
        \State Recall Top($x/2$) chunks by $sim(s, r_i)$ assign to \textbf{B}
        \State Initialize $score$ dictionary for documents
        \For{each $pair<keywords, doc>$ in list $\{A, B\}$}
            \For{each $keyword_q$ in $K$}
                \If{$keyword_q$ in $keywords$}
                    \State $score[doc]$ += 1
                \EndIf
            \EndFor
        \EndFor
        \State \textbf{ReRank} by $cross\-att$ \& $score$  
        \State \textbf{return} list $\{d_i\}$
    \EndProcedure
    
    \end{algorithmic}
    \end{algorithm}

    \subsubsection{Urban planning Knowledge Activation} \label{pka-section}

  \begin{figure}[t]
          \centering
          \includegraphics[width=0.5\textwidth]{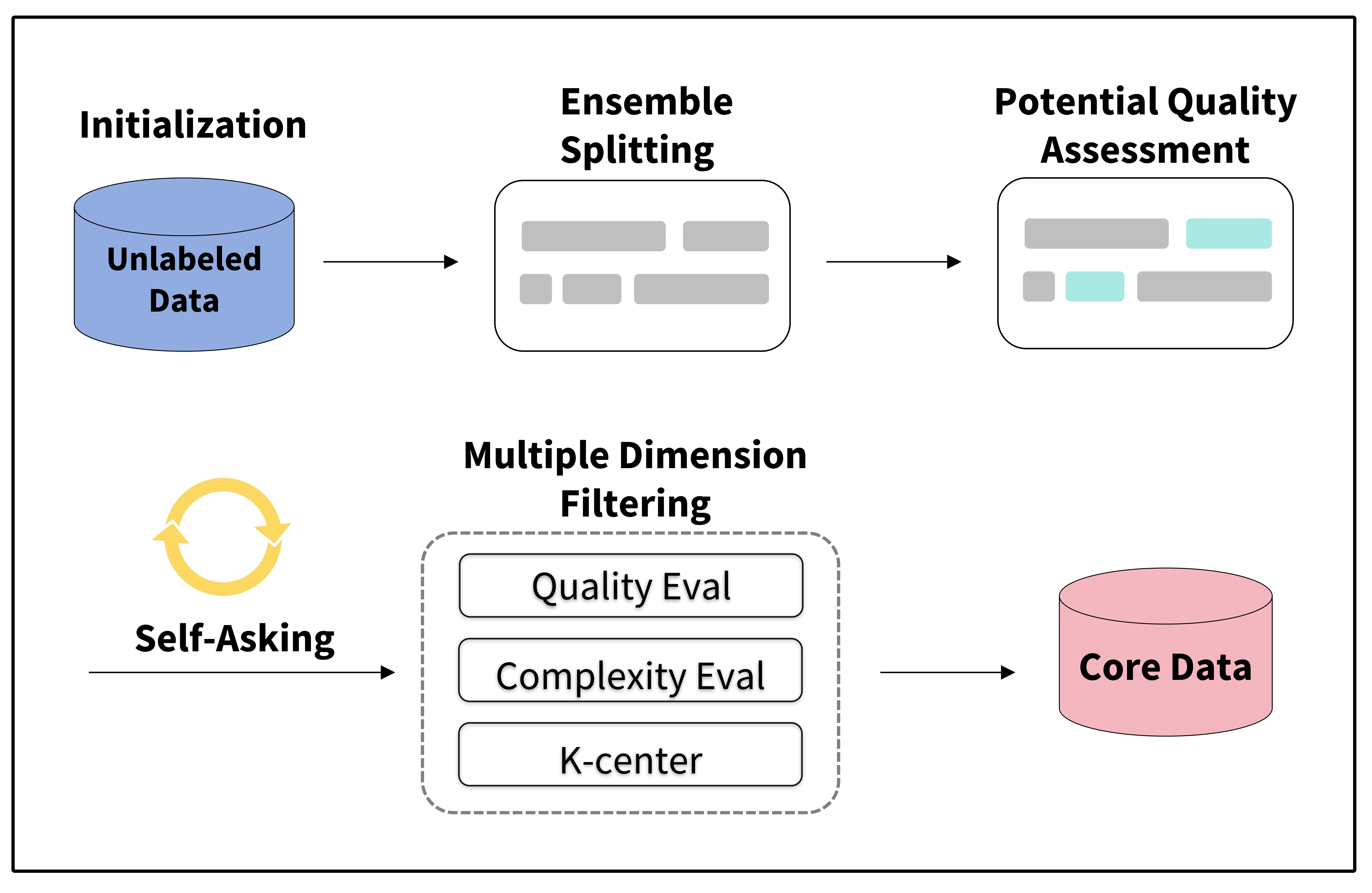}
           \caption{Urban planning-annotation}
          \label{ppl}
    \end{figure}

    Motivated by the \textbf{Humpback}\cite{li2023self} method, we propose a self-annotation technique tailored to urban planning, henceforth referred to as \textbf{Urban planning-annotation}, as illustrated in Figure \ref{ppl}. The method unfolds as follows:
  
    \begin{enumerate}
      \item \textbf{Initialization of Unlabeled Data}:
        Textual data sourced from urban planning repositories, web archives, and knowledge graphs undergo quality checks, deduplication, and sampling to produce high-quality unlabeled textual data  \( D \).
      
      \item \textbf{Ensemble Splitting}:
       We segment the unlabeled data \( D \) into multiple segments \( S_i \) using varying window sizes \( i \) and an overlap \( \Delta \) between adjacent segments, to ensure a balance between information integrity and granularity.

      \item \textbf{Potential Quality Assessment}:
        We autonomously train a scoring model to evaluate the potential urban planning knowledge value of each segment, selecting segments with higher scores as candidate segments \( P_i \).
        
    \item \textbf{Self-Asking with Random Labels}: 
        Motivated by \textbf{WizardLM}\cite{xu2023wizardlm}, we employ randomized labels across different dimensions and types to prompt large models to generate instruction \( I_i \)  deemed to possess knowledge value based on \( P_i \). Unlabeled text \( P_i \) is then either directly responded to as an answer or used to generate responses based on document-query pairs, resulting in $\langle \text{instruction}, \text{input}, \text{output} \rangle$ pairs.

      \item \textbf{Multi-dimensional Filtering}:
        The generated instructions are refined through multi-dimensional filtering, which includes instruction deduplication, quality, complexity and diversity filtration. To gauge quality and complexity, a reward model is fine-tuned leveraging sparse annotations. Taking cues from methodologies like \textbf{LIMA}\cite{zhou2024lima} and \textbf{MoDS}\cite{du2023mods}, the \textbf{k-center}\cite{sener2017active} algorithm is employed to bolster diversity in the generated instructions.
        
    \end{enumerate}

    We refer to the finely-grained data obtained through these five steps as \textit{core} data and utilize it to fine-tune base models, thus activating knowledge relevant to urban planning.

    \subsubsection{Specific Capability Development}  \label{scd-section}
    Engagement with urban planning departments and institutes reveals that large models can aid planners in generating sections for proposals, transferring styles, evaluating proposals and extracting information, but base models' limited instruction following capabilities mean prompt learning alone is insufficient to address these tasks effectively. To address practical needs in the field, we further collected over 4,000 historical versions of official plans from provinces, cities, districts, and counties nationwide for targeted capability development. We selected segments with potential utility from them and constructed self-annotated pipelines for four tasks. For example, in text style transfer, we prompt the model to simplify or colloquialize corresponding segments, then have the model rewrite them to match the desired style, generating instructions pairs t$\langle \textit{raw text}, \textit{response} \rangle$. We then employed prompt learning with varying temperatures or different models to generate responses of different quality, implementing automatic annotation to score the levels for fine-tuning the scoring model.

\subsection{PlanAgent}

    In the field of urban planning, professionals are required to have a solid grasp of domain-specific knowledge while also being proficient in utilizing tools relevant to the field. Drawing inspiration from previous work involving agents \cite{xagent2023,xie2023openagents,lagent2023,hong2023metagpt,nakajimababyagi,Significant_Gravitas_AutoGPT,wu2023autogen,MobilityAgent2023}, we have designed and developed an agent that aligns closely with the tasks and requirements of urban planning. This agent, coined as the "\textbf{PlanAgent}", is intricately tailored to suit the intricacies of urban planning endeavors.

    \subsubsection{Autonomous Todo List Generation}
    To assist urban planning professionals in executing complex tasks such as text review, audit, or evaluation, \textbf{PlanAgent} autonomously generates and optimizes task lists based on inputs from planners, subsequently executing them in sequence.
    
    \subsubsection{Orienteering Web Search} 
    \textbf{PlanAgent} utilizes \textbf{Web LLM} to access real-time planning regulations and updates. Drawing inspiration from WebGLM's web crawling \cite{webglm}, it employs vector queries and URL crawlers to ensure precision. To further enhance search accuracy, we implemented orienting URL crawlers specifically designed to identify information sources related to urban planning. 

    \subsubsection{Professional Tool Invocation}
     \textbf{PlanAgent} proficiently utilizes specialized domain-specific models to execute pivotal tasks integral to urban planning. These tasks include reverse geocoding, knowledge graph construction, and image captioning. Furthermore, PlanAgent integrates advanced tools developed by urban planning researchers for tasks such as spatiotemporal analysis\cite{liu2023social, zhang2023spatiotemporal}, transit-oriented development (TOD) settings\cite{shao2020threshold}, neighborhood life-circle urban planning\cite{zhang2022incorporating}, integrated land use and transport planning\cite{shao2023built}, urban simulations\cite{zhang2020measuring}, digital-twin city platforms, and other essential components of smart city initiatives. This holistic approach ensures a scholarly and comprehensive engagement with the intricate challenges inherent in urban planning endeavors.

    \subsubsection{Information Integration and Alignment}
    \textbf{PlanAgent} autonomously consolidates outputs from diverse LLMs (e.g., Vector LLM, Local LLM) and specialized models through advanced techniques. It can employs a customized reward model in DPO \cite{rafailov2024direct} or RLHF \cite{RLHF} to select the optimal answer, while also utilizing a summarization model to enhance findings from multiple sources.

The overarching architecture of PlanGPT is depicted as outlined above figure  \ref{plangpt}, encapsulating its multifaceted capabilities.

\section{Experiment}

In this section, we demonstrate the effectiveness of our model through extensive offline experiments. 

\subsection{Experimental Setup}
\label{subsec:experimental-setup}
\subsubsection{Training corpora}
\label{subsubsec:corpora}
For urban planning knowledge activation, we curated a specialized dataset for urban planning from diverse sources, including study materials, highly-rated Q\&A threads from urban planning forums, high-quality textbooks in related majors, and official documents published by local governments in recent years. Detailed statistics are provided in the appendix \ref{data}. Following meticulous selection using \textbf{Urban-planning-annotation}, we curated nearly 50k high-quality instruction pairs from the corpus, incorporating part of general-domain fine-tuning datasets like ShareGPT\cite{vicuna2023} or Alpaca-52k\footnote{Chinese and English Version }\cite{alpaca}, which were then used to fine-tune the base model, enhancing its urban planning abilities. 
For the development of specific capabilities, we employ urban planning data and self-annotation as detailed in Section \ref{scd-section} to generate a dataset for downstream tasks, as illustrated in Table \ref{tab:task-statistic}. Taking inspiration from LIMA, we have once again shown that even a small amount of fine-tuning data can yield satisfactory results, albeit with some instability\footnote{In practical terms, approximately 10k fine-tuning data are required to attain greater stability in outcomes.}.

\begin{table}[tb]
\small
\caption{Statistics of downstream tasks dataset. ``\#'' indicates the number of samples.}

\begin{tabular}{lcccc}
\hline
\multicolumn{1}{c}{\multirow{2}{*}{TASK}} & \multicolumn{3}{c}{\#} & \multirow{2}{*}{Metric} \\ \cline{2-4}
\multicolumn{1}{c}{}                 & Train  & Dev   & Test  &                         \\ \hline
Generating                           & 1,089  & 100 & 100  & Score                        \\
Style Transfer                       & 1,181  & 489 & 489  & Score                       \\
Information Extraction               & 1242   & 138 & 138  & Acc                        \\
Text Evaluation                      & 2345   & 100 & 100  & Acc, F1                    \\ \hline
\end{tabular}
\label{tab:task-statistic}
\vspace{-1.5em}
\end{table}

\subsubsection{Downstream Tasks}
\label{subsubsec:downstran-task}
The downstream tasks are described as follows:

\textbf{\emph{Text Generation} }
Large language models offer significant advantages in generating urban planning documentation, including comprehensive land use plans, development proposals, and zoning ordinances. By leveraging these models, urban planning professionals can streamline the process of drafting complex documents, ensuring clarity, coherence, and adherence to legal and regulatory frameworks.
To evaluate the quality of the generated content, we created a grading system from 0 to 3, with four levels indicating quality from poor to excellent. Four professional urban planners provided subjective assessments, and their average rating determined the final quality score (Human) of each model, which was then converted to a 100-point scale.

\textbf{\emph{Text Style Transfer }}
Urban planners commonly employ text style transfer techniques in their workflow. Large language models can assist in transforming brief or informal texts into the specific style of urban planning communication, thereby enhancing the efficiency of urban and rural workers. The evaluation method is similarly to \textbf{Text Generation}.

\textbf{\emph{Text Information Extraction}}
Large language models can extract key information from various textual sources, including urban planning reports, public comments, and academic studies, to support data-driven decision-making in urban and spatial planning. We self-annotate the top 5 crucial keywords for each test case and calculate accuracy (Acc), which means whether our model can predict the same keywords as we expected within an acceptable range of semantic variation.

\textbf{\emph{Text Evaluation} }
LLMs can aid urban planners in evaluating urban planning proposals by assessing the feasibility, sustainability, and community impact of diverse projects, thereby offering objective evaluations and recommendations. Notably, we simplify the evaluation process by assigning style ratings from 0 to 3 to each paragraph, treating it as a classification task with accuracy (Acc) and F1 scores. Additionally, we utilize the trained model to automatically evaluate two tasks \footnote{Text Generation, Text Style Transfer} and report the scores(PlanEval).

\begin{table*}[tb]
\small
\caption{Common Urban Planing Task Evaluation}

\resizebox{\linewidth}{!}{
\begin{tabular}{lccccccccccc}
\toprule
\multirow{2}{*}{Models} & \multicolumn{2}{c}{Generating}             & \multicolumn{2}{c}{Style Transfer} & \multicolumn{1}{c}{Information Extraction} & \multicolumn{2}{c}{Text Evaluation} & \multirow{2}{*}{Average} \\ \cmidrule(l){2-3} \cmidrule(l){4-5} \cmidrule(l){6-6}  \cmidrule(l){7-8}
                        & PlanEval  & \multicolumn{1}{l}{Human} & PlanEval  & Human & Acc  & Acc & F1    &                          \\ \midrule
ChatGLM \cite{GLM}                   & 47.67 & 41.33    & 63.94 & 67.00   &  50.00 &   26.00  & 25.67 &   54.33                  \\
\href{https://github.com/01-ai/Yi}{Yi-6B}              & 16.00 & 9.00                      & 15.41 & 12.00   &  --\footnotemark  &    20.00 & 8.33  & 10.5 \\
Baichuan2-13b-Chat\cite{baichuan2}                  & 62.67 &  34.00   & 43.90 &  39.33  &  50.32     & 33.00 &   17.42 &  36.67\\          
ChatGPT \cite{chatgpt}                  & \textbf{74.67} &  58.0                     & 66.12 & 70.67  &  --\footnotemark &  31.00  & 21.30 & 64.34                   \\
ChatGLM-2-Shots  \cite{GLM}                & 65.33 & 52.33                       & \textbf{71.10} & 63.67  & 53.81  &   30.00  & 21.76  &      58    \\           \midrule
PlanGPT $_{q}$          & 60.33 & \textbf{86.67}                       & 66.80 & \textbf{80.00}   & \textbf{65.18}  &   \textbf{41.00}  & \textbf{35.28} &   \textbf{83.34}          \\ \bottomrule
\end{tabular}
}
\label{tab:4tasks}
\end{table*}

\footnotetext{ Yi-6B only completes 10.8 \% of our tests, with the majority producing responses that do not meet our requirements.}
\footnotetext{We utilized ChatGPT \& GPT-4 for annotating the test data, therefore we are not reporting this experiment.}

\subsubsection{Baselines}
We select several baseline models for comparison:
\begin{itemize}
    \item \textbf{ChatGLM3-6B}\cite{GLM}: This is the base model of the ChatGLM3-6B series, known for its smooth dialogue and low deployment threshold.
    \item \textbf{Yi-6B}: \href{https://github.com/01-ai/Yi}{Yi-6B} is part of the Yi series, trained on a 3T multilingual corpus, showcasing strong language understanding and reasoning capabilities.
    \item \textbf{Qwen-7B}: \href{https://tongyi.aliyun.com}{Qwen-7B} is a  member of the Qwen series, featuring strong base language models pretrained on up to 2.4 trillion tokens of multilingual data with competitive performance.
    \item \href{https://platform.openai.com/docs/models/gpt-3-5-turbo} {\textbf{GPT-3.5-Turbo}}: An advanced version of GPT-3, incorporating enhancements in model size, training data, and performance across various language tasks.
    \item \textbf{Baichuan2-13B}: The \href{https://www.baichuan-ai.com/home}{Baichuan2} series introduces large-scale open-source language models, with Baichuan2-13B trained on a high-quality corpus containing 2.6 trillion tokens, showcasing top performance.
    \item \textbf{GPT4}\cite{GPT4}: The latest iteration of the Generative Pre-trained Transformer developed by OpenAI, representing a significant advancement in natural language processing technology.
 
\end{itemize}

\subsubsection{Implementation Details}
\label{subsec:impl-details}
We conduct fine-tuning experiments using four models: ChatGLM, LLaMA-chinese-7b\cite{Chinese-LLaMA-Alpaca}, Mistral-chinese-7b\cite{Chinese-Mixtral-8x7B}, and Baichuan2-13b. Eventually, we select glm3-base as our pretraining model, recognized as the state-of-the-art Chinese BaseLM with a smaller parameter scale.

Our implementation is built upon the Transformers framework \cite{wolf-etal-2020-transformers} using \href{https://pytorch.org/}{PyTorch}. For experiments involving the Local-LLM introduced in Section \ref{localllm}, we employ full-parameter fine-tuning with AdamW \cite{loshchilov2019decoupled} as the optimizer. The learning rate is initialized at 5e-5 and gradually decreased in a cosine-wise manner during training. Additionally, we utilize DeepSpeed ZeRO\-3 \cite{deepspeed} with offload and FlashAttention\-2 \cite{flashattention} to optimize memory usage, employing bfloat16 precision, with a total batch of 64.

In experiments related to PlanEmb, we also utilize AdamW as the optimizer, setting the initial learning rates to 5e-5 for pre-training and 1e-5 for fine-tuning, with a progressive decrease in learning rates as training progresses. To expedite output, we employ vllm\cite{kwon2023efficient}. with a temperature ($\tau$) of 0.95 and a top\_p value of 0.9.
Training these models typically requires about 16 hours on 8 NVIDIA 4090 GPUs.

\subsection{Offline Results}
\label{subsec:main-results}

\subsubsection{Common Urban Planing Task Evaluation}
For the aforementioned tasks, we selected prominent chat models with high rankings on the \textbf{ceval}\cite{huang2023ceval} and \textbf{cmmlu}\cite{li2023cmmlu} leaderboards to conduct experiments under zero-shot or few-shot conditions. The experimental results, along with corresponding evaluation metrics, are documented in Table \ref{tab:4tasks}. Among the four tasks, PlanGPT significantly outperformed all other models of similar scale, including proprietary models like ChatGPT, aligning closely with the awareness of urban planners. With an average 79\% Spearman correlation coefficient to human assessment, PlanEval reflects PlanGPT's effectiveness in evaluating text. However, it still faces challenges in making nuanced distinctions, such as between "best" and "good" quality. 

Furthermore, we demonstrate the model's performance during the question-answering process.

\textbf{(1) Why not use larger-scale models, such as 33b or above?}
Experimental results demonstrate that fine-tuning smaller models can achieve satisfactory results for planners. Considering the limited budgets and hardware configurations of urban planning institutes in various regions, we believe carefully tailored smaller models are competent enough for specific tasks in the urban planning domain.

\textbf{(2) Why not use prompt-learning to accomplish tasks?}
 After meticulously crafting prompts for the ChatGLM3 and comparing its performance under 2-shot conditions with that of PlanGPT under 0-shot conditions, the experimental results still indicate a noticeable gap in human evaluations. In some tasks, the performance of ChatGLM3-2-shot is even worse than under 0-shot conditions. We posit that the observed discrepancy can be ascribed to two principal factors. Initially, the limitations inherent in the instruction-following capabilities of LLMs may impede their comprehension and execution of intricate tasks. Subsequently, the constraints imposed by context length may obstruct the model's capacity to discern shared characteristics within urban planning texts. Moreover, planners may struggle with complex prompt designs, affecting work efficiency.

\textbf{(3) Why not use advanced models such as GPT, Gemini, Claude2, or models from online platforms like GLM4, Kimi-chat, etc.? } The urban planning domain exhibits extremely strong data privacy concerns closely associated with the government. Urban planning agencies prioritize data security within the institute to prevent data leaks.

\subsubsection{Urban planning Knowledge Assessment}

To ensure fairness and comprehensiveness, we utilized the \textit{urban\_and\_rural\_planner\_test} in \textbf{C-Eval}\cite{huang2023ceval}, referred to as \textbf{v1}, comprising 418 questions. \textbf{C-Eval} is recognized as a reputable Chinese evaluation suite for foundation models, featuring 13,948 multiple-choice questions across 52 diverse disciplines and four difficulty levels.

Additionally, for a broader assessment of model urban planning capabilities, we manually curated approximately 3.5k evaluation questions, including authentic questions from urban and rural planning examinations over the past decade, forming \textit{urban\_and\_rural\_planner\_test v2}. We calculated the score ratio between the two assessments, denoted as $\delta$, in which higher values indicate a more honest assessment of the model's capabilities. Notably, we strictly followed prompt templates recommended by \textit{lm-harness-test}\cite{eval-harness} and \textbf{C-Eval}, selecting options with the highest probabilities. Employing a 0-shot setting, we systematically tested models of comparable sizes listed on the leaderboard and reported their scores, as illustrated in Table \ref{tab:knowledge}.

After fine-tuning with the \textit{core} dataset as introduce in section \ref{pka-section}, our model achieved state-of-the-art performance among open-source models of similar sizes. It exhibited an approximately 5\% increase in accuracy compared to the base model. Furthermore, approaching a $\delta$ value close to 0.8 indicates the honesty and domain-generalization capabilities of our model.

\begin{table}[t]
\caption{Urban Planning Knowledge Assessment}
\centering
\begin{tabular}{lllll}
\toprule
Model                & v1    & v2    & Average & $\delta$ \\
\midrule
GPT4                & 63.2                       & \cellcolor{red!25}55.3     & \cellcolor{red!25}59.25    & \cellcolor{red!25}0.875  \\
chatgpt             & 52.2                       & 42.0                       & 47.10                      & 0.805 \\
ChatGLM3-6B         & 56.5                       & 48.8                       & 52.65                      & \cellcolor{yellow!25}0.864 \\
BlueLM-7B           & \cellcolor{yellow!25}73.0  & 27.2                       & 50.10                      & 0.373 \\
Yi-6B               & \cellcolor{red!25}73.1     & 31.2                       & 52.15                      & 0.427 \\
Baichuan-13b        & 50.5                       & 24.7                       & 37.60                      & 0.489 \\
PlanGPT\footnote{Local LLM}             
                    & 63.0                       & \cellcolor{yellow!25}51.2  & \cellcolor{yellow!25}57.10 & 0.812 \\
\bottomrule
\end{tabular}
\label{tab:knowledge}

\end{table}

\subsubsection{Assessing Plan-Emb's Proficiency}
To evaluate the performance of \textbf{Plan-Emb} in expressing specialized terminologies and language systems in urban planning, we employed the method described in Section \ref{planemb-section} to generate the \textit{urban-rural-STS-B-test} (\textbf{URSTS-B}), which consists of two levels: 0, indicating no relation, and 1, signifying a stronger correlation between the word and its explanation. We rigorously evaluated the performance of various phases of \textbf{Plan-Emb} on URSTS-B and other general datasets, employing Spearman's correlation coefficient \cite{spearman1961proof} for assessment. As shown in the table \ref{tab:embedding result}, it's obvious that with the help of the fine-tuning stage, Plan-Emb holds more information in urban planning than any general models, which indicates that our embedding strategy exhibits superior aggregational efficacy. Furthermore, it is noteworthy that as training progresses, BERT-cse significantly outperforms BERT-base, underscoring the critical importance of the first-stage pretrain.

\begin{table}[tb]
\centering
\caption{Embedding Performance Comparison\footnotemark}

\resizebox{\linewidth}{!}{

\begin{tabular}{llllll}
\toprule
             & URSTS-B & STS-B\footnotemark & PAWSX\footnotemark   & LCQMC\footnotemark    & AVG  \\
\midrule
BERT    & 0.176& 0.186& 0.085& 0.038& 0.121\\
BERT+     & 0.503& 0.400& 0.057& 0.234& 0.299\\
Roberta & 0.176& 0.277& 0.113& 0.228& 0.199\\
Roberta+   & 0.594& 0.530& 0.147& 0.512& 0.446\\
BERT-cse     & 0.739&\textbf{0.772}& 0.126& \textbf{0.704}& 0.585\\
Plan-Emb     & \textbf{0.754}& 0.753& \textbf{0.157}& 0.693& \textbf{0.589}\\

\bottomrule
\end{tabular}
}
\label{tab:embedding result}
\end{table}

\footnotetext{\textit{+} denotes the model's performance after the initial pretraining stage using SBERT with a portion of the training data, while BERT-cse reflects the model's performance after being fully pretrained with CoSENT.}

\footnotetext{STS-B\cite{cer-etal-2017-semeval}}
\footnotetext{PAWSX\cite{DBLP:journals/corr/abs-1908-11828}}
\footnotetext{LCQMC\cite{liu-etal-2018-lcqmc}}

\begin{figure}
     \centering
     \includegraphics[width=1\linewidth]{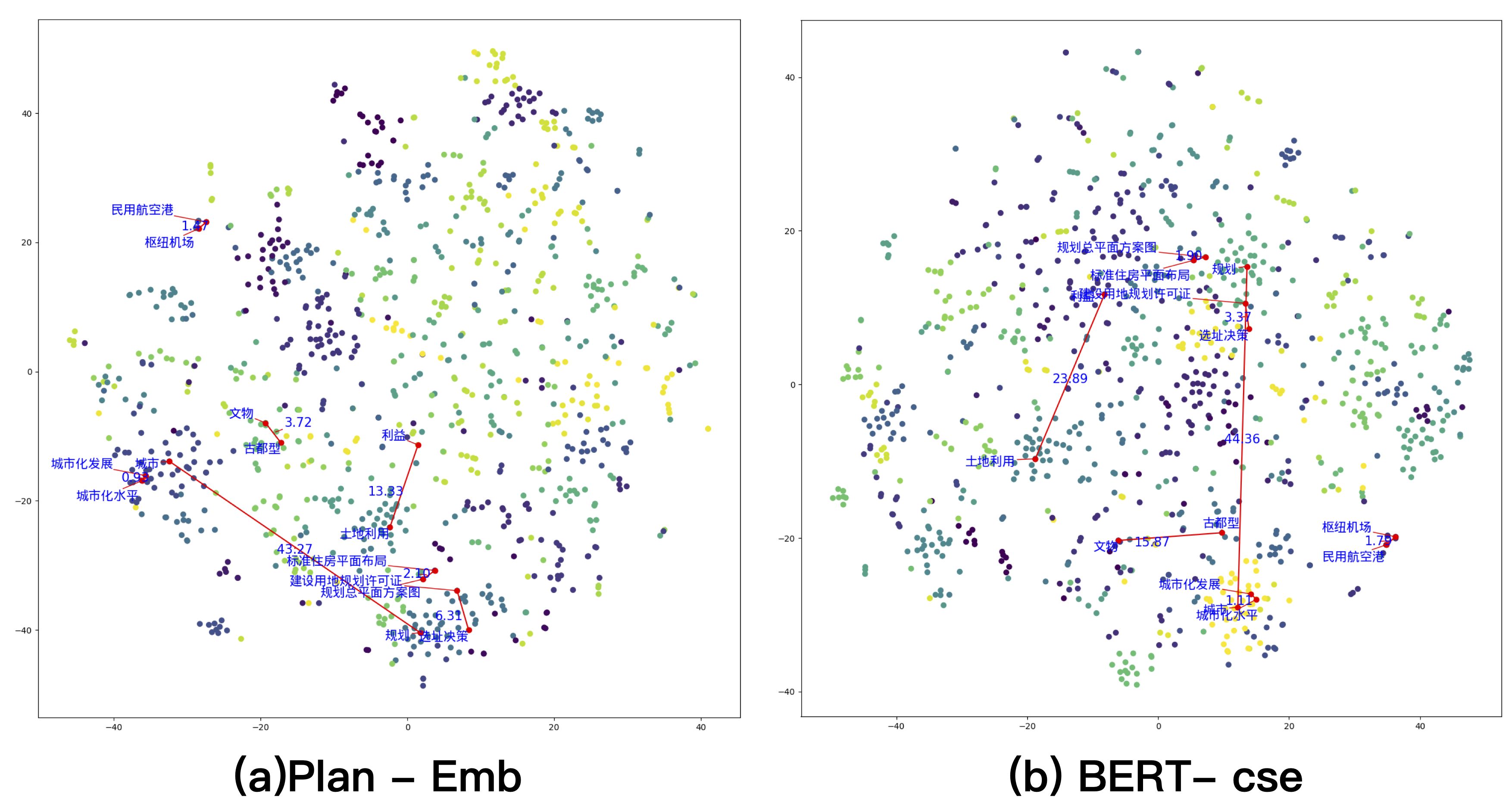}
     \caption{The \textbf{t-SNE} projection between Plan-Emb and BERT-cse.}
     \label{fig:Plan-Emb}
\end{figure}

A visualization of the t-SNE\cite{t-SNE} projection between Plan-Emb and BERT-cse is shown in \ref{fig:Plan-Emb}. From the marked examples, we can draw the conclusion that Plan-Emb learns the relation in urban and rural planning much better than BERT-cse in most cases. The terms \textit{land utilization} ("\textit{\begin{CJK*}{UTF8}{gbsn}土地利用\end{CJK*}}") and \textit{benefits} ("\textit{\begin{CJK*}{UTF8}{gbsn}利益\end{CJK*}}"), along with those representing \textit{ancient capital type} ("\textit{\begin{CJK*}{UTF8}{gbsn}古都型\end{CJK*}}") and \textit{cultural relics} ("\textit{\begin{CJK*}{UTF8}{gbsn}文物\end{CJK*}}"), which frequently co-occur in urban planning documents, exhibit significantly reduced distances in the t-SNE projection space of Plan-Emb compared to BERT-cse. Additionally, \textit{standard residential floor plan layout}, \textit{construction land planning permit}, and \textit{planned total area schematic diagram}, all indicative of domain knowledge in regional planning, demonstrate enhanced aggregative properties within Plan-Emb.

\begin{table}[tb]
\caption{Ablation Studies for Vector-LLM}
\resizebox{\linewidth}{!}{
\begin{tabular}{lcccl}
\toprule
Method                & score@1    & score@5    & AVG   \\
\midrule
Raw Search            & 48.7  & 48.5  & 48.6  \\
Raw Search + PlanEmb  & 49.7  & 48.8  & 49.3  \\
VectorLLM (all) & \textbf{51.9}  & \textbf{52.4}  & \textbf{52.2}  \\

\bottomrule
\end{tabular}
}
\label{tab:hs}
\end{table}

\subsubsection{Ablation Studies for \textbf{Vector-LLM}}
Ablation experiments were conducted on \textbf{Vector-LLM} to demonstrate the effectiveness of customized modules in enhancing downstream task performance. Following the design of previous experimental settings, we extracted appropriate segments from a large corpus of text to answer questions in \textit{urban\_and\_rural\_planner\_test}, and calculated \textit{score@k}, representing the accuracy of answered questions within the top \textit{k} segments. To ensure fairness, network retrieval tools were disabled, and model judgments were based solely on contextual and intrinsic knowledge. We systematically removed \textbf{Plan-Emb} and \textbf{Plan-HS}, documenting the experimental outcomes in Table \ref{tab:hs}. Our findings indicate that the removal of any task component led to a decline in performance. Specifically, the elimination of each component (\textbf{Plan-Emb} and \textbf{Plan-HS}) resulted in score reductions of 0.7\% and 3.6\%, respectively.
This indirectly highlights the superior expressive capability of \textbf{Plan-Emb} for urban planning texts. Additionally, it's worth noting that \textbf{Plan-HS} effectively tackled issues related to texts with a low signal-to-noise ratio, significantly enhancing information utilization and accuracy.

\subsection{Case Study}
\label{subsec:task-study}

In this section, we will discuss relevant tasks in the domain of real-world urban planning and provide potential solutions.

\subsubsection{TASK: Review}
Review is the primary task of urban planning institute staff, as extensively discussed in Section \ref{introduction}, which consumes a significant amount of time. By utilizing VectorLLM to identify reference standard to document queries and then conducting reviews using PlanAgent, we believe that LLMs can detect inconsistencies, inaccuracies, or discrepancies within the text, ensuring the integrity and quality of urban planning proposals.

However, in practical work, we have found that despite sophisticated prompting, large models often fail to align with human consciousness, exhibiting extremes by either detecting minor errors that could be overlooked or excessively relaxing standards, resulting in lower recall rates.

Our solution involves employing GPT-4 to randomly introduce partial errors into urban planning text, along with indicating their locations. Our staff then identify error reasons, categorized into three types: 1. factual errors 2. spelling/grammar errors 3. stylistic errors (including harmful language). Initially, we refine the cognitive capabilities of large-scale models to discern the mere presence of errors. Subsequently, we instruct them to identify and flag errors.

\subsubsection{TASK: Evaluation}

In the urban planning domain, text evaluation is a complex task, including verifying the framework of the text, reviewing the details and style of the text (as in the aforementioned review steps), and scoring the overall nature of the document. The overall nature of the document includes novelty, feasibility, and utility.

\begin{enumerate}
    \item \textbf{Novelty}: Assessing the differences and connections with historical urban planning.
    \item \textbf{Feasibility}: Urban planning needs to consider comprehensive conditions such as local economic level, geographical conditions, and interpersonal relationships.
    \item \textbf{Utility}: Whether the urban planning can solve practical problems.
\end{enumerate}
In actual operations, our solutions are as follows:
\textbf{Novelty}: We will use vectorLLM to quickly retrieve and match historical urban planning.
\textbf{Feasibility}: \textbf{PlanAgent} integrates network search tools and multimodal capabilities to solve.
\textbf{Utility}: To evaluate the efficacy of the proposed plan, we will develop a simulation environment where multiple \textbf{PlanAgent}s will engage in role-playing activities. Through simulated interactions and scenario analyses, the plan's effectiveness will be assessed across diverse contexts.

\begin{figure}[t]
  \centering
  \includegraphics[width=0.5\textwidth]{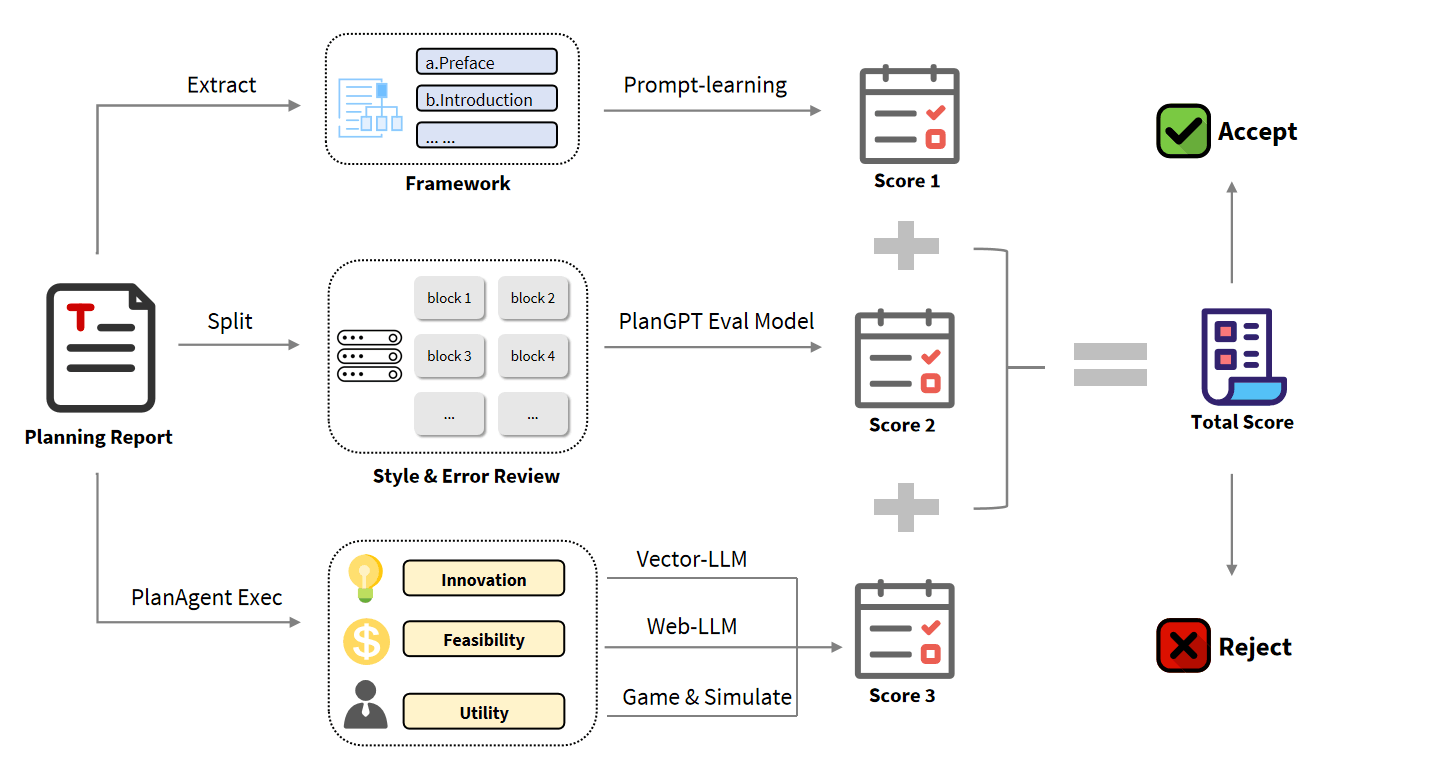}
  \caption{Assessment Task process}
  \label{usage}
\end{figure}

\section{Future Work}
In our future endeavors, we aim to explore several key directions to further the advancement of urban and spatial planning:

\begin{itemize}
\item \textbf{PlanGPT model refinement:} We will expand our ongoing efforts in large-scale model pre-training, specifically focusing on urban planning. Our goal is to enrich the knowledge base for both urban and rural planning contexts.

\item \textbf{Utilization of Multi-Modal Techniques:} We will investigate the application of multimodal techniques within urban planning to achieve a more comprehensive understanding of spatial dynamics. Currently, we are actively developing \textbf{PlanVLM} for this purpose.

\item \textbf{Gradual Integration of AI Solutions:} Our strategy involves deep collaboration with urban planning institutions to better understand practical needs, refine workflows involving large models, and address procedural challenges in urban planning effectively.
\end{itemize}

We advocate for a comprehensive overhaul of future urban planning frameworks. By addressing industry concerns and promoting progressive strategies, we envision a gradual yet impactful transformation of future urban planning practices.

\section{Conclusion}
In this Paper, we introduced PlanGPT, the first large-scale language model framework designed specifically for the field of urban and spatial planning. Through a customized approach, we successfully addressed challenges in urban planning text management, review, and assessment, demonstrating its efficiency and superiority in practice. Our work signifies a significant step forward in the convergence of artificial intelligence and urban and rural planning, providing planners with powerful support tools and facilitating more intelligent and efficient decision-making in urban and rural development. In the future, we will continue to refine and expand the capabilities of PlanGPT to further advance its application in the urban planning domain.

\section{Acknowledgements} 
We would like to thank other members of the BSAI Lab, for their help in collecting and cleaning data, along with the Natural Language Processing Research Group at Southern University of Science and Technology. This work was supported by funding from the National Natural Science Foundation of China (42171201) and the Shenzhen Municipal Natural Science Foundation (Key Project) (GXWD20221231165807007-20200810223326001).  PlanGPT has been further promoted and applied by various organizations including the Spatial Planning Bureau of the Ministry of Natural Resources of China, the China Land Surveying and Planning Institute, the Planning and Natural Resources Bureau of Shenzhen Municipality, the Planning and Research Center of Guangzhou Municipality, the Shenzhen Marine Development Promotion Research Center, the China Academy of Urban Planning and Design, and Guangzhou Planning Corporation.
\clearpage
\bibliography{custom}

\begin{thebibliography}{83}
\expandafter\ifx\csname natexlab\endcsname\relax\def\natexlab#1{#1}\fi

\bibitem[{Aminabadi et~al.(2022)Aminabadi, Rajbhandari, Awan, Li, Li, Zheng,
  Ruwase, Smith, Zhang, Rasley et~al.}]{deepspeed}
Reza~Yazdani Aminabadi, Samyam Rajbhandari, Ammar~Ahmad Awan, Cheng Li, Du~Li,
  Elton Zheng, Olatunji Ruwase, Shaden Smith, Minjia Zhang, Jeff Rasley, et~al.
  2022.
\newblock Deepspeed-inference: enabling efficient inference of transformer
  models at unprecedented scale.
\newblock In \emph{SC22: International Conference for High Performance
  Computing, Networking, Storage and Analysis}, pages 1--15. IEEE.

\bibitem[{Anthropic(2023)}]{Claude2}
Anthropic. 2023.
\newblock Model card and evaluations for claude models.

\bibitem[{Asai et~al.(2023)Asai, Wu, Wang, Sil, and Hajishirzi}]{asai2023self}
Akari Asai, Zeqiu Wu, Yizhong Wang, Avirup Sil, and Hannaneh Hajishirzi. 2023.
\newblock Self-rag: Learning to retrieve, generate, and critique through
  self-reflection.
\newblock \emph{arXiv preprint arXiv:2310.11511}.

\bibitem[{Baichuan(2023)}]{baichuan2}
Baichuan. 2023.
\newblock \href {https://arxiv.org/abs/2309.10305} {Baichuan 2: Open
  large-scale language models}.
\newblock \emph{arXiv preprint arXiv:2309.10305}.

\bibitem[{Borgeaud et~al.(2022)Borgeaud, Mensch, Hoffmann, Cai, Rutherford,
  Millican, Van Den~Driessche, Lespiau, Damoc, Clark
  et~al.}]{borgeaud2022improving}
Sebastian Borgeaud, Arthur Mensch, Jordan Hoffmann, Trevor Cai, Eliza
  Rutherford, Katie Millican, George~Bm Van Den~Driessche, Jean-Baptiste
  Lespiau, Bogdan Damoc, Aidan Clark, et~al. 2022.
\newblock Improving language models by retrieving from trillions of tokens.
\newblock In \emph{International conference on machine learning}, pages
  2206--2240. PMLR.

\bibitem[{Bowman et~al.(2015)Bowman, Angeli, Potts, and
  Manning}]{snli:emnlp2015}
Samuel~R. Bowman, Gabor Angeli, Christopher Potts, and Christopher~D. Manning.
  2015.
\newblock A large annotated corpus for learning natural language inference.
\newblock In \emph{Proceedings of the 2015 Conference on Empirical Methods in
  Natural Language Processing (EMNLP)}. Association for Computational
  Linguistics.

\bibitem[{Cer et~al.(2017)Cer, Diab, Agirre, Lopez-Gazpio, and
  Specia}]{cer-etal-2017-semeval}
Daniel Cer, Mona Diab, Eneko Agirre, I{\~n}igo Lopez-Gazpio, and Lucia Specia.
  2017.
\newblock \href {https://doi.org/10.18653/v1/S17-2001} {{S}em{E}val-2017 task
  1: Semantic textual similarity multilingual and crosslingual focused
  evaluation}.
\newblock In \emph{Proceedings of the 11th International Workshop on Semantic
  Evaluation ({S}em{E}val-2017)}, pages 1--14, Vancouver, Canada. Association
  for Computational Linguistics.

\bibitem[{Chen et~al.(2024)Chen, Xiao, Zhang, Luo, Lian, and Liu}]{chen2024bge}
Jianlv Chen, Shitao Xiao, Peitian Zhang, Kun Luo, Defu Lian, and Zheng Liu.
  2024.
\newblock \href {http://arxiv.org/abs/2402.03216} {Bge m3-embedding:
  Multi-lingual, multi-functionality, multi-granularity text embeddings through
  self-knowledge distillation}.

\bibitem[{Cheng et~al.(2024)Cheng, Luo, Chen, Liu, Zhao, and
  Yan}]{cheng2024lift}
Xin Cheng, Di~Luo, Xiuying Chen, Lemao Liu, Dongyan Zhao, and Rui Yan. 2024.
\newblock Lift yourself up: Retrieval-augmented text generation with
  self-memory.
\newblock \emph{Advances in Neural Information Processing Systems}, 36.

\bibitem[{Chiang et~al.(2023)Chiang, Li, Lin, Sheng, Wu, Zhang, Zheng, Zhuang,
  Zhuang, Gonzalez, Stoica, and Xing}]{vicuna2023}
Wei-Lin Chiang, Zhuohan Li, Zi~Lin, Ying Sheng, Zhanghao Wu, Hao Zhang, Lianmin
  Zheng, Siyuan Zhuang, Yonghao Zhuang, Joseph~E. Gonzalez, Ion Stoica, and
  Eric~P. Xing. 2023.
\newblock \href {https://lmsys.org/blog/2023-03-30-vicuna/} {Vicuna: An
  open-source chatbot impressing gpt-4 with 90\%* chatgpt quality}.

\bibitem[{Christiano et~al.(2017)Christiano, Leike, Brown, Martic, Legg, and
  Amodei}]{RLHF}
Paul Christiano, Jan Leike, Tom~B. Brown, Miljan Martic, Shane Legg, and Dario
  Amodei. 2017.
\newblock \href {http://arxiv.org/abs/arXiv:1706.03741} {Deep reinforcement
  learning from human preferences}.

\bibitem[{Cui et~al.(2023{\natexlab{a}})Cui, Li, Yan, Chen, and Yuan}]{ChatLaw}
Jiaxi Cui, Zongjian Li, Yang Yan, Bohua Chen, and Li~Yuan. 2023{\natexlab{a}}.
\newblock Chatlaw.
\newblock \url{https://github.com/PKU-YuanGroup/ChatLaw}.

\bibitem[{Cui et~al.(2021)Cui, Che, Liu, Qin, and
  Yang}]{cui-etal-2021-pretrain}
Yiming Cui, Wanxiang Che, Ting Liu, Bing Qin, and Ziqing Yang. 2021.
\newblock \href {https://doi.org/10.1109/TASLP.2021.3124365} {Pre-training with
  whole word masking for chinese bert}.

\bibitem[{Cui et~al.(2022)Cui, Che, Wang, and Liu}]{cui2022lert}
Yiming Cui, Wanxiang Che, Shijin Wang, and Ting Liu. 2022.
\newblock \href {http://arxiv.org/abs/2211.05344} {Lert: A
  linguistically-motivated pre-trained language model}.

\bibitem[{Cui et~al.(2023{\natexlab{b}})Cui, Yang, and
  Yao}]{Chinese-LLaMA-Alpaca}
Yiming Cui, Ziqing Yang, and Xin Yao. 2023{\natexlab{b}}.
\newblock \href {https://arxiv.org/abs/2304.08177} {Efficient and effective
  text encoding for chinese llama and alpaca}.
\newblock \emph{arXiv preprint arXiv:2304.08177}.

\bibitem[{Dao(2023)}]{flashattention}
Tri Dao. 2023.
\newblock Flashattention-2: Faster attention with better parallelism and work
  partitioning.
\newblock \emph{arXiv preprint arXiv:2307.08691}.

\bibitem[{DeepMind(2023{\natexlab{a}})}]{bard}
Google DeepMind. 2023{\natexlab{a}}.
\newblock Bard.
\newblock \url{https://bard.google.com}.

\bibitem[{DeepMind(2023{\natexlab{b}})}]{gemini}
Google DeepMind. 2023{\natexlab{b}}.
\newblock Gemini.
\newblock \url{https://gemini.google.com}.

\bibitem[{Du et~al.(2023)Du, Zong, and Zhang}]{du2023mods}
Qianlong Du, Chengqing Zong, and Jiajun Zhang. 2023.
\newblock Mods: Model-oriented data selection for instruction tuning.
\newblock \emph{arXiv preprint arXiv:2311.15653}.

\bibitem[{Du et~al.(2022)Du, Qian, Liu, Ding, Qiu, Yang, and Tang}]{GLM}
Zhengxiao Du, Yujie Qian, Xiao Liu, Ming Ding, Jiezhong Qiu, Zhilin Yang, and
  Jie Tang. 2022.
\newblock Glm: General language model pretraining with autoregressive blank
  infilling.
\newblock In \emph{Proceedings of the 60th Annual Meeting of the Association
  for Computational Linguistics (Volume 1: Long Papers)}, pages 320--335.

\bibitem[{et~al.(2023{\natexlab{a}})}]{Prithvi-100M}
Jakubik et~al. 2023{\natexlab{a}}.
\newblock \href {https://doi.org/10.57967/hf/0952} {{Prithvi-100M}}.

\bibitem[{et~al.(2023{\natexlab{b}})}]{PaLM2}
Rohan~Anil et~al. 2023{\natexlab{b}}.
\newblock \href {http://arxiv.org/abs/arXiv:2305.10403} {Palm 2 technical
  report}.

\bibitem[{Gao et~al.(2023)Gao, Tow, Abbasi, Biderman, Black, DiPofi, Foster,
  Golding, Hsu, Le~Noac'h, Li, McDonell, Muennighoff, Ociepa, Phang, Reynolds,
  Schoelkopf, Skowron, Sutawika, Tang, Thite, Wang, Wang, and
  Zou}]{eval-harness}
Leo Gao, Jonathan Tow, Baber Abbasi, Stella Biderman, Sid Black, Anthony
  DiPofi, Charles Foster, Laurence Golding, Jeffrey Hsu, Alain Le~Noac'h,
  Haonan Li, Kyle McDonell, Niklas Muennighoff, Chris Ociepa, Jason Phang,
  Laria Reynolds, Hailey Schoelkopf, Aviya Skowron, Lintang Sutawika, Eric
  Tang, Anish Thite, Ben Wang, Kevin Wang, and Andy Zou. 2023.
\newblock \href {https://doi.org/10.5281/zenodo.10256836} {A framework for
  few-shot language model evaluation}.

\bibitem[{Gao et~al.(2022)Gao, Yao, and Chen}]{gao2022simcse}
Tianyu Gao, Xingcheng Yao, and Danqi Chen. 2022.
\newblock \href {http://arxiv.org/abs/2104.08821} {Simcse: Simple contrastive
  learning of sentence embeddings}.

\bibitem[{HIT-SCIR(2024)}]{Chinese-Mixtral-8x7B}
HIT-SCIR. 2024.
\newblock Chinese-mixtral-8x7b: An open-source mixture-of-experts llm.
\newblock \url{https://github.com/HIT-SCIR/Chinese-Mixtral-8x7B}.

\bibitem[{Hong et~al.(2023)Hong, Zheng, Chen, Cheng, Wang, Zhang, Wang, Yau,
  Lin, Zhou et~al.}]{hong2023metagpt}
Sirui Hong, Xiawu Zheng, Jonathan Chen, Yuheng Cheng, Jinlin Wang, Ceyao Zhang,
  Zili Wang, Steven Ka~Shing Yau, Zijuan Lin, Liyang Zhou, et~al. 2023.
\newblock Metagpt: Meta programming for multi-agent collaborative framework.
\newblock \emph{arXiv preprint arXiv:2308.00352}.

\bibitem[{Huang et~al.(2023{\natexlab{a}})Huang, Ping, Xu, Shoeybi, Chang, and
  Catanzaro}]{huang2023raven}
Jie Huang, Wei Ping, Peng Xu, Mohammad Shoeybi, Kevin Chen-Chuan Chang, and
  Bryan Catanzaro. 2023{\natexlab{a}}.
\newblock Raven: In-context learning with retrieval augmented encoder-decoder
  language models.
\newblock \emph{arXiv preprint arXiv:2308.07922}.

\bibitem[{Huang et~al.(2023{\natexlab{b}})Huang, Bai, Zhu, Zhang, Zhang, Su,
  Liu, Lv, Zhang, Lei, Fu, Sun, and He}]{huang2023ceval}
Yuzhen Huang, Yuzhuo Bai, Zhihao Zhu, Junlei Zhang, Jinghan Zhang, Tangjun Su,
  Junteng Liu, Chuancheng Lv, Yikai Zhang, Jiayi Lei, Yao Fu, Maosong Sun, and
  Junxian He. 2023{\natexlab{b}}.
\newblock C-eval: A multi-level multi-discipline chinese evaluation suite for
  foundation models.
\newblock In \emph{Advances in Neural Information Processing Systems}.

\bibitem[{Jiang et~al.(2023)Jiang, Xu, Gao, Sun, Liu, Dwivedi-Yu, Yang, Callan,
  and Neubig}]{jiang2023active}
Zhengbao Jiang, Frank~F Xu, Luyu Gao, Zhiqing Sun, Qian Liu, Jane Dwivedi-Yu,
  Yiming Yang, Jamie Callan, and Graham Neubig. 2023.
\newblock Active retrieval augmented generation.
\newblock \emph{arXiv preprint arXiv:2305.06983}.

\bibitem[{Kim et~al.(2023)Kim, Kim, Jeon, Park, and Kang}]{kim2023tree}
Gangwoo Kim, Sungdong Kim, Byeongguk Jeon, Joonsuk Park, and Jaewoo Kang. 2023.
\newblock Tree of clarifications: Answering ambiguous questions with
  retrieval-augmented large language models.
\newblock \emph{arXiv preprint arXiv:2310.14696}.

\bibitem[{Kwon et~al.(2023)Kwon, Li, Zhuang, Sheng, Zheng, Yu, Gonzalez, Zhang,
  and Stoica}]{kwon2023efficient}
Woosuk Kwon, Zhuohan Li, Siyuan Zhuang, Ying Sheng, Lianmin Zheng, Cody~Hao Yu,
  Joseph~E. Gonzalez, Hao Zhang, and Ion Stoica. 2023.
\newblock \href {http://arxiv.org/abs/2309.06180} {Efficient memory management
  for large language model serving with pagedattention}.

\bibitem[{Li et~al.(2023{\natexlab{a}})Li, Zhang, Koto, Yang, Zhao, Gong, Duan,
  and Baldwin}]{li2023cmmlu}
Haonan Li, Yixuan Zhang, Fajri Koto, Yifei Yang, Hai Zhao, Yeyun Gong, Nan
  Duan, and Timothy Baldwin. 2023{\natexlab{a}}.
\newblock \href {http://arxiv.org/abs/2306.09212} {Cmmlu: Measuring massive
  multitask language understanding in chinese}.

\bibitem[{Li et~al.(2023{\natexlab{b}})Li, Zhang, Li, Chen, Chen, Cheng, Wang,
  Zhou, and Xiao}]{Li2023FromQT}
Ming Li, Yong Zhang, Zhitao Li, Jiuhai Chen, Lichang Chen, Ning Cheng, Jianzong
  Wang, Tianyi Zhou, and Jing Xiao. 2023{\natexlab{b}}.
\newblock \href {https://api.semanticscholar.org/CorpusID:261076515} {From
  quantity to quality: Boosting llm performance with self-guided data selection
  for instruction tuning}.
\newblock \emph{ArXiv}, abs/2308.12032.

\bibitem[{Li et~al.(2023{\natexlab{c}})Li, Yu, Zhou, Schick, Zettlemoyer, Levy,
  Weston, and Lewis}]{li2023self}
Xian Li, Ping Yu, Chunting Zhou, Timo Schick, Luke Zettlemoyer, Omer Levy,
  Jason Weston, and Mike Lewis. 2023{\natexlab{c}}.
\newblock Self-alignment with instruction backtranslation.
\newblock \emph{arXiv preprint arXiv:2308.06259}.

\bibitem[{Lin et~al.(2023)Lin, Chen, Chen, Shi, Lomeli, James, Rodriguez, Kahn,
  Szilvasy, Lewis et~al.}]{lin2023ra}
Xi~Victoria Lin, Xilun Chen, Mingda Chen, Weijia Shi, Maria Lomeli, Rich James,
  Pedro Rodriguez, Jacob Kahn, Gergely Szilvasy, Mike Lewis, et~al. 2023.
\newblock Ra-dit: Retrieval-augmented dual instruction tuning.
\newblock \emph{arXiv preprint arXiv:2310.01352}.

\bibitem[{Liu and Zhang(2023)}]{liu2023social}
C.~Liu and W.~Zhang. 2023.
\newblock Social and spatial heterogeneities in covid-19 impacts on
  individual's metro use: A big-data driven causality inference.
\newblock \emph{Applied Geography}, 155:102947.

\bibitem[{Liu et~al.(2024)Liu, Zeng, He, Jiang, and He}]{liu2024what}
Wei Liu, Weihao Zeng, Keqing He, Yong Jiang, and Junxian He. 2024.
\newblock \href {https://openreview.net/forum?id=BTKAeLqLMw} {What makes good
  data for alignment? a comprehensive study of automatic data selection in
  instruction tuning}.
\newblock In \emph{The Twelfth International Conference on Learning
  Representations}.

\bibitem[{Liu et~al.(2023)Liu, Lai, Yu, Xu, Zeng, Du, Zhang, Dong, and
  Tang}]{webglm}
Xiao Liu, Hanyu Lai, Hao Yu, Yifan Xu, Aohan Zeng, Zhengxiao Du, Peng Zhang,
  Yuxiao Dong, and Jie Tang. 2023.
\newblock \href {http://arxiv.org/abs/2306.07906} {Webglm: Towards an efficient
  web-enhanced question answering system with human preferences}.

\bibitem[{Liu et~al.(2018)Liu, Chen, Deng, Zeng, Chen, Li, and
  Tang}]{liu-etal-2018-lcqmc}
Xin Liu, Qingcai Chen, Chong Deng, Huajun Zeng, Jing Chen, Dongfang Li, and
  Buzhou Tang. 2018.
\newblock \href {https://aclanthology.org/C18-1166} {{LCQMC}:a large-scale
  {C}hinese question matching corpus}.
\newblock In \emph{Proceedings of the 27th International Conference on
  Computational Linguistics}, pages 1952--1962, Santa Fe, New Mexico, USA.
  Association for Computational Linguistics.

\bibitem[{Loshchilov and Hutter(2019)}]{loshchilov2019decoupled}
Ilya Loshchilov and Frank Hutter. 2019.
\newblock \href {http://arxiv.org/abs/1711.05101} {Decoupled weight decay
  regularization}.

\bibitem[{Lou et~al.(2023)Lou, Zhang, Xie, Sun, Ahn, Xu, Su, and
  Yin}]{lou2023muffin}
Renze Lou, Kai Zhang, Jian Xie, Yuxuan Sun, Janice Ahn, Hanzi Xu, Yu~Su, and
  Wenpeng Yin. 2023.
\newblock Muffin: Curating multi-faceted instructions for improving instruction
  following.
\newblock In \emph{The Twelfth International Conference on Learning
  Representations}.

\bibitem[{Lun et~al.(2023)Lun, Tao, Su, Zhu, and Fan}]{MobilityAgent2023}
Tianle Lun, Yicheng Tao, Junyou Su, He~Zhu, and Zipei Fan. 2023.
\newblock Mobilityagent.
\newblock \url{https://github.com/XiaoLeGG/mobility-agent}.

\bibitem[{Mikolov et~al.(2013)Mikolov, Chen, Corrado, and
  Dean}]{mikolov2013efficient}
Tomas Mikolov, Kai Chen, Greg Corrado, and Jeffrey Dean. 2013.
\newblock \href {http://arxiv.org/abs/1301.3781} {Efficient estimation of word
  representations in vector space}.

\bibitem[{Mistral-AI(2023)}]{mistral}
Mistral-AI. 2023.
\newblock mistral.
\newblock \url{https://mistral.ai/}.

\bibitem[{Nakajima()}]{nakajimababyagi}
Yohei Nakajima.
\newblock Babyagi, 2023.
\newblock \emph{URL https://github. com/yoheinakajima/babyagi. GitHub
  repository}.

\bibitem[{Nakano et~al.(2021)Nakano, Hilton, Balaji, Wu, Ouyang, Kim, Hesse,
  Jain, Kosaraju, Saunders et~al.}]{nakano2021webgpt}
Reiichiro Nakano, Jacob Hilton, Suchir Balaji, Jeff Wu, Long Ouyang, Christina
  Kim, Christopher Hesse, Shantanu Jain, Vineet Kosaraju, William Saunders,
  et~al. 2021.
\newblock Webgpt: Browser-assisted question-answering with human feedback.
\newblock \emph{arXiv preprint arXiv:2112.09332}.

\bibitem[{Oord et~al.(2018)Oord, Li, and Vinyals}]{oord2018representation}
Aaron van~den Oord, Yazhe Li, and Oriol Vinyals. 2018.
\newblock Representation learning with contrastive predictive coding.
\newblock \emph{arXiv preprint arXiv:1807.03748}.

\bibitem[{OpenAI(2022)}]{chatgpt}
OpenAI. 2022.
\newblock Chatgpt.
\newblock \url{https://chat.openai.com}.

\bibitem[{OpenAI(2023)}]{GPT4}
OpenAI. 2023.
\newblock \href {http://arxiv.org/abs/arXiv:2303.08774} {Gpt-4 technical
  report}.

\bibitem[{Peng(2023)}]{TransGPT}
Wang Peng. 2023.
\newblock \href {https://github.com/DUOMO/TransGPT} {Duomo/transgpt}.

\bibitem[{Rafailov et~al.(2024)Rafailov, Sharma, Mitchell, Manning, Ermon, and
  Finn}]{rafailov2024direct}
Rafael Rafailov, Archit Sharma, Eric Mitchell, Christopher~D Manning, Stefano
  Ermon, and Chelsea Finn. 2024.
\newblock Direct preference optimization: Your language model is secretly a
  reward model.
\newblock \emph{Advances in Neural Information Processing Systems}, 36.

\bibitem[{Reimers and Gurevych(2019)}]{reimers2019sentencebert}
Nils Reimers and Iryna Gurevych. 2019.
\newblock \href {http://arxiv.org/abs/1908.10084} {Sentence-bert: Sentence
  embeddings using siamese bert-networks}.

\bibitem[{Sener and Savarese(2017)}]{sener2017active}
Ozan Sener and Silvio Savarese. 2017.
\newblock Active learning for convolutional neural networks: A core-set
  approach.
\newblock \emph{arXiv preprint arXiv:1708.00489}.

\bibitem[{Shao et~al.(2020)Shao, Zhang, Cao, Yang, and Yin}]{shao2020threshold}
Q.~Shao, W.~Zhang, X.~Cao, J.~Yang, and J.~Yin. 2020.
\newblock Threshold and moderating effects of land use on metro ridership in
  shenzhen: Implications for tod planning.
\newblock \emph{Journal of Transport Geography}, 89:102878.

\bibitem[{Shao et~al.(2023)Shao, Zhang, Cao, and Yang}]{shao2023built}
Q.~Shao, W.~Zhang, X.~J. Cao, and J.~Yang. 2023.
\newblock Built environment interventions for emission mitigation: A machine
  learning analysis of travel-related co2 in a developing city.
\newblock \emph{Journal of Transport Geography}, 110:103632.

\bibitem[{{Significant Gravitas}()}]{Significant_Gravitas_AutoGPT}
{Significant Gravitas}.
\newblock \href {https://github.com/Significant-Gravitas/AutoGPT} {{AutoGPT}}.

\bibitem[{Spearman(1961)}]{spearman1961proof}
Charles Spearman. 1961.
\newblock The proof and measurement of association between two things.

\bibitem[{Su(2022)}]{kexuefm-8847}
Jianlin Su. 2022.
\newblock Cosent.

\bibitem[{Sun et~al.(2022)Sun, Wang, Tay, Yang, and Zhou}]{sun2022recitation}
Zhiqing Sun, Xuezhi Wang, Yi~Tay, Yiming Yang, and Denny Zhou. 2022.
\newblock Recitation-augmented language models.
\newblock \emph{arXiv preprint arXiv:2210.01296}.

\bibitem[{Taori et~al.(2023)Taori, Gulrajani, Zhang, Dubois, Li, Guestrin,
  Liang, and Hashimoto}]{alpaca}
Rohan Taori, Ishaan Gulrajani, Tianyi Zhang, Yann Dubois, Xuechen Li, Carlos
  Guestrin, Percy Liang, and Tatsunori~B. Hashimoto. 2023.
\newblock Stanford alpaca: An instruction-following llama model.
\newblock \url{https://github.com/tatsu-lab/stanford_alpaca}.

\bibitem[{Team(2023{\natexlab{a}})}]{lagent2023}
Lagent~Developer Team. 2023{\natexlab{a}}.
\newblock {Lagent: InternLM} a lightweight open-source framework that allows
  users to efficiently build large language model(llm)-based agents.
\newblock \url{https://github.com/InternLM/lagent}.

\bibitem[{Team(2023{\natexlab{b}})}]{xagent2023}
XAgent Team. 2023{\natexlab{b}}.
\newblock Xagent: An autonomous agent for complex task solving.

\bibitem[{Touvron et~al.(2023)Touvron, Lavril, Izacard, Martinet, Lachaux,
  Lacroix, Rozi{\`e}re, Goyal, Hambro, Azhar et~al.}]{touvron2023llama}
Hugo Touvron, Thibaut Lavril, Gautier Izacard, Xavier Martinet, Marie-Anne
  Lachaux, Timoth{\'e}e Lacroix, Baptiste Rozi{\`e}re, Naman Goyal, Eric
  Hambro, Faisal Azhar, et~al. 2023.
\newblock Llama: Open and efficient foundation language models.
\newblock \emph{arXiv preprint arXiv:2302.13971}.

\bibitem[{Tycho~Young(2023)}]{MathGPT}
Krish~Mangroila Tycho~Young, Andy~Zhang. 2023.
\newblock Mathgpt - an exploration into the field of mathematics with large
  language models.

\bibitem[{Van~der Maaten and Hinton(2008)}]{t-SNE}
Laurens Van~der Maaten and Geoffrey Hinton. 2008.
\newblock Visualizing data using t-sne.
\newblock \emph{Journal of machine learning research}, 9(11).

\bibitem[{Wang et~al.(2023)Wang, Liu, Xi, Qiang, Zhao, Qin, and Liu}]{huatuo}
Haochun Wang, Chi Liu, Nuwa Xi, Zewen Qiang, Sendong Zhao, Bing Qin, and Ting
  Liu. 2023.
\newblock Huatuo: Tuning llama model with chinese medical knowledge.
\newblock \emph{arXiv preprint arXiv:2304.06975}.

\bibitem[{Wang et~al.(2022)Wang, Kordi, Mishra, Liu, Smith, Khashabi, and
  Hajishirzi}]{wang2022self}
Yizhong Wang, Yeganeh Kordi, Swaroop Mishra, Alisa Liu, Noah~A Smith, Daniel
  Khashabi, and Hannaneh Hajishirzi. 2022.
\newblock Self-instruct: Aligning language model with self generated
  instructions.
\newblock \emph{arXiv preprint arXiv:2212.10560}.

\bibitem[{Wei et~al.(2023)Wei, Zhao, Zhang, Zhu, Wang, Yang, Li, Cheng, Lü,
  Hu, Li, Yang, Luo, Wu, Liu, Cheng, Cheng, Zhang, Zhang, Lin, Wang, Ma, Dong,
  Sun, Chen, Peng, Liang, Yan, Fang, and Zhou}]{wei2023skywork}
Tianwen Wei, Liang Zhao, Lichang Zhang, Bo~Zhu, Lijie Wang, Haihua Yang, Biye
  Li, Cheng Cheng, Weiwei Lü, Rui Hu, Chenxia Li, Liu Yang, Xilin Luo, Xuejie
  Wu, Lunan Liu, Wenjun Cheng, Peng Cheng, Jianhao Zhang, Xiaoyu Zhang, Lei
  Lin, Xiaokun Wang, Yutuan Ma, Chuanhai Dong, Yanqi Sun, Yifu Chen, Yongyi
  Peng, Xiaojuan Liang, Shuicheng Yan, Han Fang, and Yahui Zhou. 2023.
\newblock \href {http://arxiv.org/abs/2310.19341} {Skywork: A more open
  bilingual foundation model}.

\bibitem[{Wolf et~al.(2020)Wolf, Debut, Sanh, Chaumond, Delangue, Moi, Cistac,
  Rault, Louf, Funtowicz, Davison, Shleifer, von Platen, Ma, Jernite, Plu, Xu,
  Le~Scao, Gugger, Drame, Lhoest, and Rush}]{wolf-etal-2020-transformers}
Thomas Wolf, Lysandre Debut, Victor Sanh, Julien Chaumond, Clement Delangue,
  Anthony Moi, Pierric Cistac, Tim Rault, Remi Louf, Morgan Funtowicz, Joe
  Davison, Sam Shleifer, Patrick von Platen, Clara Ma, Yacine Jernite, Julien
  Plu, Canwen Xu, Teven Le~Scao, Sylvain Gugger, Mariama Drame, Quentin Lhoest,
  and Alexander Rush. 2020.
\newblock \href {https://doi.org/10.18653/v1/2020.emnlp-demos.6} {Transformers:
  State-of-the-art natural language processing}.
\newblock In \emph{Proceedings of the 2020 Conference on Empirical Methods in
  Natural Language Processing: System Demonstrations}, pages 38--45, Online.
  Association for Computational Linguistics.

\bibitem[{Wu et~al.(2023)Wu, Bansal, Zhang, Wu, Li, Zhu, Jiang, Zhang, Zhang,
  Liu, Awadallah, White, Burger, and Wang}]{wu2023autogen}
Qingyun Wu, Gagan Bansal, Jieyu Zhang, Yiran Wu, Beibin Li, Erkang Zhu,
  Li~Jiang, Xiaoyun Zhang, Shaokun Zhang, Jiale Liu, Ahmed~Hassan Awadallah,
  Ryen~W White, Doug Burger, and Chi Wang. 2023.
\newblock \href {http://arxiv.org/abs/2308.08155} {Autogen: Enabling next-gen
  llm applications via multi-agent conversation framework}.

\bibitem[{Xie et~al.(2023)Xie, Zhou, Cheng, Shi, Weng, Liu, Hua, Zhao, Liu, Liu
  et~al.}]{xie2023openagents}
Tianbao Xie, Fan Zhou, Zhoujun Cheng, Peng Shi, Luoxuan Weng, Yitao Liu,
  Toh~Jing Hua, Junning Zhao, Qian Liu, Che Liu, et~al. 2023.
\newblock Openagents: An open platform for language agents in the wild.
\newblock \emph{arXiv preprint arXiv:2310.10634}.

\bibitem[{Xiong et~al.(2023)Xiong, Wang, Zhu, Zhao, Liu, Wang, and
  Shen}]{Doctorglm}
Honglin Xiong, Sheng Wang, Yitao Zhu, Zihao Zhao, Yuxiao Liu, Qian Wang, and
  Dinggang Shen. 2023.
\newblock Doctorglm: Fine-tuning your chinese doctor is not a herculean task.
\newblock \emph{arXiv preprint arXiv:2304.01097}.

\bibitem[{Xu et~al.(2023)Xu, Sun, Zheng, Geng, Zhao, Feng, Tao, and
  Jiang}]{xu2023wizardlm}
Can Xu, Qingfeng Sun, Kai Zheng, Xiubo Geng, Pu~Zhao, Jiazhan Feng, Chongyang
  Tao, and Daxin Jiang. 2023.
\newblock Wizardlm: Empowering large language models to follow complex
  instructions.
\newblock \emph{arXiv preprint arXiv:2304.12244}.

\bibitem[{Yang et~al.(2019)Yang, Zhang, Tar, and
  Baldridge}]{DBLP:journals/corr/abs-1908-11828}
Yinfei Yang, Yuan Zhang, Chris Tar, and Jason Baldridge. 2019.
\newblock \href {http://arxiv.org/abs/1908.11828} {{PAWS-X:} {A} cross-lingual
  adversarial dataset for paraphrase identification}.
\newblock \emph{CoRR}, abs/1908.11828.

\bibitem[{Yu et~al.(2022)Yu, Iter, Wang, Xu, Ju, Sanyal, Zhu, Zeng, and
  Jiang}]{yu2022generate}
Wenhao Yu, Dan Iter, Shuohang Wang, Yichong Xu, Mingxuan Ju, Soumya Sanyal,
  Chenguang Zhu, Michael Zeng, and Meng Jiang. 2022.
\newblock Generate rather than retrieve: Large language models are strong
  context generators.
\newblock \emph{arXiv preprint arXiv:2209.10063}.

\bibitem[{Zhang et~al.(2023{\natexlab{a}})Zhang, Fu, Zhang, Yu, and
  Cai}]{TrafficGPT}
Siyao Zhang, Daocheng Fu, Zhao Zhang, Bin Yu, and Pinlong Cai.
  2023{\natexlab{a}}.
\newblock \href {http://arxiv.org/abs/arXiv:2309.06719} {Trafficgpt: Viewing,
  processing and interacting with traffic foundation models}.

\bibitem[{Zhang et~al.(2020)Zhang, Fang, Zhou, and Zhu}]{zhang2020measuring}
W.~Zhang, C.~Fang, L.~Zhou, and J.~Zhu. 2020.
\newblock Measuring megaregional structure in the pearl river delta by mobile
  phone signaling data: A complex network approach.
\newblock \emph{Cities}, 104:102809.

\bibitem[{Zhang et~al.(2022)Zhang, Lu, Zhao, Luo, and
  Yin}]{zhang2022incorporating}
W.~Zhang, D.~Lu, Y.~Zhao, X.~Luo, and J.~Yin. 2022.
\newblock Incorporating polycentric development and neighborhood life-circle
  planning for reducing driving in beijing: Nonlinear and threshold analysis.
\newblock \emph{Cities}, 121:103488.

\bibitem[{Zhang and Ning(2023)}]{zhang2023spatiotemporal}
W.~Zhang and K.~Ning. 2023.
\newblock Spatiotemporal heterogeneities in the causal effects of mobility
  intervention policies during the covid-19 outbreak: A spatially interrupted
  time-series (sits) analysis.
\newblock \emph{Annals of the American Association of Geographers},
  113(5):1112--1134.

\bibitem[{Zhang et~al.(2024)Zhang, Cai, Zhang, Zhuang, and Mao}]{earthgpt}
Wei Zhang, Miaoxin Cai, Tong Zhang, Yin Zhuang, and Xuerui Mao. 2024.
\newblock Earthgpt: A universal multi-modal large language model for
  multi-sensor image comprehension in remote sensing domain.
\newblock \emph{arXiv preprint arXiv:2401.16822}.

\bibitem[{Zhang et~al.(2023{\natexlab{b}})Zhang, Yang, and Xu}]{XuanYuan}
Xuanyu Zhang, Qing Yang, and Dongliang Xu. 2023{\natexlab{b}}.
\newblock \href {http://arxiv.org/abs/arXiv:2305.12002} {Xuanyuan 2.0: A large
  chinese financial chat model with hundreds of billions parameters}.

\bibitem[{Zheng et~al.(2024)Zheng, Guo, Qu, Guo, Zhang, Du, Lin, Huang, Chen,
  Fu et~al.}]{zheng2024kun}
Tianyu Zheng, Shuyue Guo, Xingwei Qu, Jiawei Guo, Weixu Zhang, Xinrun Du,
  Chenghua Lin, Wenhao Huang, Wenhu Chen, Jie Fu, et~al. 2024.
\newblock Kun: Answer polishment for chinese self-alignment with instruction
  back-translation.
\newblock \emph{arXiv preprint arXiv:2401.06477}.

\bibitem[{Zhou et~al.(2024)Zhou, Liu, Xu, Iyer, Sun, Mao, Ma, Efrat, Yu, Yu
  et~al.}]{zhou2024lima}
Chunting Zhou, Pengfei Liu, Puxin Xu, Srinivasan Iyer, Jiao Sun, Yuning Mao,
  Xuezhe Ma, Avia Efrat, Ping Yu, Lili Yu, et~al. 2024.
\newblock Lima: Less is more for alignment.
\newblock \emph{Advances in Neural Information Processing Systems}, 36.

\end{thebibliography}
\bibliographystyle{acl_natbib}
\clearpage
\section{Appendix}
\subsection{Fine-tuning instruction pair}
\subsubsection{Generating}

\textbf{Prompt}:\\
\begin{CJK*}{UTF8}{gbsn}
假设你是一个善于写中国城乡规划文本的工作人员, 请你根据 \#城乡规划文本特征\# 对 \#原文\# 进行续写:\\
\#城乡规划文本特征\#：\\
1. 文字连贯，内容完整。\\
2. 用语规范，内容官方严谨，不使用口语化的表达方式，不出现英文、拼音、网络用语等。\\
3. 符合政府文件风格, 语言优美, 内容充实, 不出现概念堆叠与重复内容。\\
\#原文\#：
符合城乡规划风格公文有以下要求：\\1. 文字连贯，内容完整。\\2. 用语规范，不使用口语化的表达方式，不出现英文、拼音、网络用语等。\\请你帮我续写以下城市规划公文：\\\#开头\#\\定远，一座衢通九省的枢纽之城，一座历史悠久的人文之城，一座物产丰饶的产业之城。建立国土空间规划体系并监督实施是党中央全面贯彻生态文明建设的重大国策。 《定远县国土空间总体规划（2021-2035年）》\\
\#续写\#\\
\end{CJK*}
\textbf{Response}:\\
\begin{CJK*}{UTF8}{gbsn}
定远，一座衢通九省的枢纽之城，一座历史悠久的人文之城，一座物产丰饶的产业之城。建立国土空间规划体系并监督实施是党中央全面贯彻生态文明建设的重大国策。《定远县国土空间总体规划（2021-2035年）》是遵循相关法律法规，统筹考虑定远县自然环境承载力、社会经济发展需求和资源约束，对全县国土空间开发保护作出的总体安排和综合部署，是指导和开展各类国土空间开发保护活动、进行空间用途管制的基本依据，是编制下级国土空间总体规划、详细规划和相关专项规划的依据。
\end{CJK*}

\subsubsection{Style Transfer}

\textbf{Prompt}:\\
\begin{CJK*}{UTF8}{gbsn}
假设你是一个善于写中国城乡规划文本的工作人员, 请你根据 \#城乡规划文本特征\# 将以下 \#原文\# 改写成符合城乡规划风格的文字。\\
\#城乡规划文本特征\#：\\
1. 文字连贯，内容完整。\\
2. 用语规范，内容官方严谨，不使用口语化的表达方式，不出现英文、拼音、网络用语等。\\
3. 符合政府文件风格, 语言优美, 内容充实, 不出现概念堆叠与重复内容。\\
\#原文\#：\\
这份规划是庆安县国土空间的全面规划，旨在指导各类开发建设活动，保护土地资源，规划国土空间，管理空间发展。该规划是根据国家和省级工作安排，结合庆安县实际情况制定的文件。在规划中考虑了庆安县的资源环境承载能力和国土空间开发适宜性，评估了现有国土空间开发保护情况，明确了全县国土空间的资源特征和目标。我们的目标是积极落实国家、省、市级要求，与重大区域战略协同发展。规划还包括优化三类空间布局、明确控制线、划分规划分区及用途分类、加强自然和人文资源保护利用，提升国土空间品质，完善基础支撑体系，制定国土空间整治和生态修复任务，优化中心城区空间布局，明确发展方向，提出规划传导指引和近期建设计划，完善规划实施保障机制，为庆安建设更加富强、民主、和谐、美丽的社会主义现代化新庆安提供国土空间保障。\\
\#城乡规划风格文字\#\\
\end{CJK*}
\textbf{Response}:\\
\begin{CJK*}{UTF8}{gbsn}
庆安县国土空间总体规划（2021-2035年）是全县国土空间开发保护的战略性、纲领性、综合性规划，是编制各类国土空间详细规划、专项规划、专项空间的依据，是对各类国土空间开发保护活动的战略引导。根据国家和省级国土空间规划要求，结合庆安县实际，本次规划以服务和保障庆安县国土空间保护、合理开发和综合利用，促进全县经济持续发展、社会持续进步、环境持续改善和民生持续改善，为庆安县建设美丽富强、民主和谐、幸福新庆安提供规划支撑。
\end{CJK*}

\subsubsection{Text Evaluation}
\textbf{Prompt}:\\
\begin{CJK*}{UTF8}{gbsn}
根据如下 \#评分标准\# 对 \#国土空间规划文本\# 进行评分, 请只输出0到3中的一个数字。\\
\#评分标准\#：\\
0: 文本不完整、模糊、离题、混乱或者无法评分。\\
1: 文本是连贯、完整且内容与城市规划相关的，但整体用语偏口语化不规范。\\
2: 文本基本符合符合规划文件风格, 但语言空洞乏味、内容重复、用词不当。\\
3: 文本完全符合规划文件风格，语言优美、内容充实、用词准确。\\
\\
\#国土空间规划文本\#:\\
国土空间规划是国家空间发展的指南、可持续发展的空间蓝图，是各类开发保护建设活动的基本依据。建立国土空间规划体系并监督实施，将主体功能区规划、土地利用规划、城乡规划等空间规划融合为统一的国土空间规划，实现“多规合一”，强化国土空间规划对各专项规划的指导约束作用，是党中央、国务院作出的重大部署。根据《中共中央国务院关于建立国土空间规划体系并监督实施的若干意见》同江西省及九江市的统一部署，德安县组织编制了《德安县国土空间总体规划(2021-2035)》(以下简称《规划》）。《规划》是德安县域国土空间保护、开发、利用、修复的行动纲领，是编制详细规划、相关专项规划、乡镇国土空间总体规划，以及开展各类开发保护建设活动、实施国土空间用途管制基本依据。规划目标年为2035年，近期至2025年，远景展望至2050年。\\
分数：\\
\end{CJK*}
\textbf{Response}:\\
3

\subsubsection{Information Extraction}
\textbf{Prompt}:\\
\begin{CJK*}{UTF8}{gbsn}
请你为以下内容写出五个关键词：\\
步入生态文明时代，党中央、国务院作出了建立国土空间规划体系并监督实施的部署，开启了“多规融合”的国土空间规划新里程。揭西县人民政府编制了《揭西县国土空间总体规划（2020-2035年）》（以下简称《规划》）。《规划》深入贯彻落实习近平总书记对广东重要讲话和重要批示指示精神,是对县域范围内国土空间开发保护做出的总体安排和综合部署，是落实新发展理念、实施高效能空间治理，促进高质量发展和高品质生活的空间政策，是一定时期内揭西国土空间保护、开发、利用、修复和指导各类建设的行动总纲，是编制下位国土空间总体规划、详细规划、相关专项规划的基本依据，具有战略性、综合性、协调性和约束性。\\
\#\#\#关键词：\#\#\#\\
\end{CJK*}
\textbf{Response}:\\
\begin{CJK*}{UTF8}{gbsn}
1.揭西县\\
2.国土空间规划\\
3.多规融合\\
4.国土空间开发保护\\
5.高质量发展\\
\end{CJK*}

\subsection{Urban and Rural Planner Test V2 Question Samples}
\begin{CJK*}{UTF8}{gbsn}
\begin{enumerate}
    \item 城市发展与社会关系错误的是\_\_\_\_。
    \begin{enumerate}
        \item 城市是社会矛盾的集合体
        \item 城市是社会问题集中发正地
        \item 城市中旧的社会问题的解决不会带来新的社会问题
        \item 社会问题的解决是城市发展目标和现实动力
    \end{enumerate}
    \textbf{Answer}: c\\
    \item 关于文艺复兴和绝对君权时期，欧洲城市建设特征的表述，正确的是\_\_\_\_。
    \begin{enumerate}
        \item 文艺复兴时期，具有古典风格的广场，街道是地市的主要特征
        \item 文艺复兴时期，众多中世纪新建成的城市进行了系统的有机更新
        \item 绝对君权时期，在欧洲国家首都建设中，伦敦城市改建影响最大
        \item 绝对君权时期，纵横交错的大道是城市建设的典型特征之一
    \end{enumerate}
    \textbf{Answer}: a
    \item 根据《市级国土空间总体规划编制指南（试行）》，居住用地规划内容要求不包括\_\_\_\_。
    \begin{enumerate}
        \item 优化空间结构和功能布局、改善职住关系
        \item 引导政策性住房优先布局在交通和就业便利地区
        \item 进一步提升人均居住用地面积
        \item 严控高层高密度住宅
    \end{enumerate}
    \textbf{Answer}: c
\end{enumerate}
\end{CJK*}

\subsection{urban-rural-STS-B-test Samples}

\begin{table}[htbp]
\centering
\begin{tabularx}{\columnwidth}{|>{\hsize=0.2\hsize}X|>{\hsize=0.6\hsize}X|>{\hsize=0.2\hsize}X|}
\hline
\textbf{Keyword}    & \textbf{Explanation} & \textbf{Rating} \\ \hline
\begin{CJK*}{UTF8}{gbsn}煤炭
\end{CJK*}   
&
\begin{CJK*}{UTF8}{gbsn}生物多样性的维护与平衡。\end{CJK*}  & 
\begin{CJK*}{UTF8}{gbsn}0\end{CJK*} \\ \hline

\begin{CJK*}{UTF8}{gbsn}水资源开发利用
\end{CJK*}              
& 
\begin{CJK*}{UTF8}{gbsn}消防队员正在救火\end{CJK*}  & 
\begin{CJK*}{UTF8}{gbsn}0\end{CJK*} \\ \hline

\begin{CJK*}{UTF8}{gbsn}产业名城	
\end{CJK*}
& 
\begin{CJK*}{UTF8}{gbsn}产业聚集的城市，以产业为主要经济支柱。\end{CJK*}
&
\begin{CJK*}{UTF8}{gbsn}1\end{CJK*} \\ \hline
\end{tabularx}
\caption{urban-rural-STS-B-test Samples}
\label{format}
\end{table}

\clearpage
\begin{table*}[htbp]
\centering
\begin{tabularx}{\textwidth}{l|X}
\hline
\textbf{Model}    & \textbf{Output Answer}   \\ \hline
PlanGPT & \begin{CJK*}{UTF8}{gbsn}大鹏新区地处深圳东部沿海地区大鹏半岛，是深圳所辖功能区。大鹏新区是深圳生态资源最为集中地区 ，亦是深圳最早有古人类活动的地区之一，拥有深圳历史遗存的宝贵生态与人文财富。《大鹏新区国土空间总体规划（2021-2035年）》是大鹏新区未来国土空间发展的蓝图和战略部署，是落实新发展理念、融入新发展格局、优化国土空间开发保护格局、提升城市品质竞争力的重要举措，是对深圳市国土空间总体规划、大鹏新区本级规划的引领和协调，是实现大鹏新区高质量发展、高品质生活、高效能治理的法定手段。规划是对区域自然地理条件、资源禀赋、产业基础、治理能力和治理需求的综合评价和总体部署，是未来十年间大鹏新区国土空间开发、保护、利用、修复的总纲领，是编制各专项专项规划、详细规划、村庄规划、相关专项规划的基础。本规划立足区域发展和战略定位，以国家战略为引领，以资源环境承载力为底线，以国土空间开发保护底线、资源约束性空间需求和战略布局为统筹，统筹构建新区国土空间开发保护格局，为深圳建设全球宜居宜业、宜山近海、宜业现代、生态美丽的高品质中国样本提供空间保障。\end{CJK*}                                         \\ \hline
ChatGLM  &  \begin{CJK*}{UTF8}{gbsn} 大鹏新区位于深圳，是生态资源最为集中的地区之一，也是深圳最早有古人类活动的地区之一。这里拥有深圳历史遗存的宝贵生态与人文财富，是深圳寻求保护与发展相协调的重要试验场。
大鹏新区是国家生态文明建设和循环经济发展的重要试验区，同时也是深圳建设全球性生态城市的窗口和平台。其自然环境优美，拥有长达13公里的海岸线，丰富的海岸带生态系统和海洋资源。此地也是深圳重要的生态保护区，保护了大量的海洋生物和栖息地。
此外，大鹏新区还拥有丰富的历史文化遗存，如大鹏所城、沙头角炮台、东门古城等，是深圳历史文化的重要代表。
为了保护和发展大鹏新区的生态资源，深圳市政府采取了一系列措施，包括加强海岸线管理、保护海洋生态系统、推动海洋经济发展和推广低碳生活方式等。同时，这里也建立了一批国家公园和自然保护区，如大鹏半岛国家地质公园、大鹏湾国家级自然保护区等，对这里的生态环境进行了有效的保护和管理。 \end{CJK*}                                      \\ \hline
Standard Answer &   \begin{CJK*}{UTF8}{gbsn} 大鹏新区位于深圳东部沿海地区，坐落于大鹏半岛，是深圳市辖下的一个功能区。该区域拥有丰富的生态资源，也是深圳历史上古人类活动的重要区域之一，保留了深圳的历史遗产，融合了宝贵的生态和人文财富。大鹏新区成为深圳寻求生态保护与发展协调发展的重要试验场，同时也是国家生态文明建设示范区和国家"绿水青山就是金山银山"实践创新基地。规划的指导思想基于习近平新时代中国特色社会主义思想，全面贯彻党的二十大精神，立足新的发展阶段，贯彻新发展理念，服务并融入新的发展格局。规划强调高质量发展，坚持生态优先和绿色发展，以人民为中心，促进人与自然的和谐共生。整体谋划了大鹏新区面向2035年中长期发展的空间战略蓝图，旨在为深圳打造"人与自然和谐共生的美丽中国典范"提供有力支撑。此规划的制定考虑了国家和地方层面的相关要求，全面贯彻国土空间规划体系以及深圳市国土空间总体规划的相关内容。它将指导和约束下层级的详细规划编制，以确保市级和区级规划之间的有效传导。在遵循资源环境承载能力的前提下，规划强调了挖掘资源潜力，全面管理和控制所有要素，加强陆海协同发展，同时平衡生态保护和经济发展，旨在探索实现人与自然和谐共生的大鹏绿色发展之路。
\end{CJK*}     \\ \hline
\end{tabularx}
\caption{Planning text generation comparison }

\label{format}
\end{table*}

\begin{table*}[htbp]
\centering
\footnotesize
\resizebox{\linewidth}{!}{
\begin{tabularx}{\textwidth}{l|X}
\hline
\textbf{Model/Subject}    & \textbf{Output Answer}   \\ \hline
% Original & \begin{CJK*}{UTF8}{gbsn}
% 即墨，一座底蕴深厚的千年商都，拥有7000多年人类活动史、2300多年建置史和1400多年建城史。即墨发展，始于战国陆通东西的商贸重镇，兴于清朝金口港海通南北的海陆枢纽，而后即墨作为青岛的下辖县市，承接青岛的功能外溢，伴随着改革开放的大潮，即墨的商贸物流产业强势发展，孕育出蓬勃发展的民营经济，逐步成为江北商贸名城，跻身全国百强县级市前列。2017年撤市设区以来，在习近平新时代中国特色社会主义思想指引下，即墨全方位推进融青发展，抢抓海洋强国、“一带一路”和乡村振兴等国家重大战略决策机遇，以国土空间规划编制为契机，全面谋划至2035年的发展新路径，积极融入胶东经济圈一体化及青岛都市区协同发展大框架，加快构建高质量发展的国土空间新格局，为实现第二个百年目标奠定良好的基础。国土空间规划编制是党中央、国务院作出的重大部署，是将主体功能区规划、土地利用规划、城乡规划等空间规划融合统一，实现“多规合一”。即墨区国土空间分区规划，是在青岛市国土空间总体规划指导下完成的空间发展指南，是即墨可持续发展的空间蓝图，是各类开发保护建设活动的基本依据。这本手册力求以简洁的图文语言，描绘即墨国土空间的发展蓝图，有助于您更好了解即墨未来的建设目标和发展方向。遇见即墨，预见2035。
% \end{CJK*}                                         \\ \hline
% Informal Style & \begin{CJK*}{UTF8}{gbsn}
% 即墨市是一个有着千年历史的商都，拥有7000多年的人类活动史、2300多年的建置史和1400多年的建城史。该城市的发展始于战国时期，是陆上贸易重镇，后来随着清朝金口港的兴盛，逐渐成为海陆交通枢纽。随着改革开放的推进，即墨作为青岛的下辖县市，接受了青岛的辐射，商贸物流产业迅速发展，民营经济也随之崛起，逐渐成为中国北方的重要商贸名城。2017年即墨撤市设区后，我们遵循习近平新时代中国特色社会主义思想，全面推动融入青岛都市区和胶东经济圈的发展，抓住国家重大战略决策机遇，全面规划到2035年的发展新路径，积极融入胶东经济圈一体化协同发展框架，构建高质量发展的新空间格局，为实现第二个百年目标奠定坚实基础。国土空间规划编制是党中央、国务院的重要部署，旨在将各种空间规划融合在一起，实现“多规合一”。即墨区国土空间分区规划是在青岛市国土空间总体规划指导下完成的空间发展指南，是即墨未来发展的空间蓝图，也是各种开发、保护和建设活动的基本依据。本手册力求以简洁的图文语言，描绘即墨国土空间的发展蓝图，让您更好地了解即墨未来的建设目标和发展方向。来即墨，共同见证2035的美好。
% \end{CJK*}                                         \\ \hline
PlanGPT & \begin{CJK*}{UTF8}{gbsn}
即墨市，这座拥有千年历史商都底蕴的城市，拥有7000多年的人类活动史、2300多年的建置史与1400多年的建城史。自战国时期发展起，即墨作为陆上贸易重镇，后因清朝金口港的兴盛，逐渐发展为海陆交通枢纽。改革开放以来，即墨作为青岛下辖县市，在青岛的辐射下，商贸物流产业迅速发展，民营经济异军突起，成为中国北方的重要商贸名城。2017年，即墨撤市设区，全面融入青岛都市区和胶东经济圈发展，遵循习近平新时代中国特色社会主义思想，推动融入胶东经济圈一体化协同发展框架，全面规划到2035年的发展新路径，积极融入胶东经济圈一体化协同发展框架，构建高质量发展的新空间格局，为实现第二个百年目标奠定坚实基础。国土空间规划编制是党中央、国务院的重要部署，旨在将各种空间规划融合在一起，实现“多规合一”。即墨区国土空间分区规划是在青岛市国土空间总体规划指导下完成的空间发展指南，是即墨未来发展的空间蓝图，也是各种开发、保护和建设活动的基本依据。本手册力求以简洁的图文语言，描绘即墨国土空间的发展蓝图，让您更好地了解即墨未来的建设目标和发展方向。欢迎您来即墨，共同见证2035的美好。
\end{CJK*}                                         \\ \hline
ChatGLM-2shot  &  \begin{CJK*}{UTF8}{gbsn}即墨市，作为一个拥有千年历史商都的城市，具有超过7000年的人类活动史、2300多年的建置史以及1400多年的建城史。城市的形成始于战国时期，曾是陆上贸易重镇。后来，随着清朝金口港的兴盛，逐渐发展为海陆交通枢纽。改革开放的推进使即墨成为青岛下辖的县市，受到青岛的辐射，商贸物流产业迅速发展，民营经济也随之崛起，使其逐渐成为中国北方的重要商贸名城。2017年，即墨撤市设区，我们遵循习近平新时代中国特色社会主义思想，全面推动融入青岛都市区和胶东经济圈的发展，抓住国家重大战略决策机遇，全面规划到2035年的发展新路径，积极融入胶东经济圈一体化协同发展框架，构建高质量发展的新空间格局，为实现第二个百年目标奠定坚实基础。国土空间规划编制是党中央、国务院的重要部署，旨在实现“多规合一”。即墨区国土空间分区规划是在青岛市国土空间总体规划指导下完成的空间发展指南，是即墨未来发展的空间蓝图，也是各种开发、保护和建设活动的基本依据。本手册以简洁的图文语言，描绘即墨国土空间的发展蓝图，让您更好地了解即墨未来的建设目标和发展方向。欢迎您来即墨，共同见证2035的美好未来。\end{CJK*}                                      \\ \hline
Baichuan2-13B-Chat  &   \begin{CJK*}{UTF8}{gbsn}
即墨区作为千年商都，拥有7000多年人类活动史、2300多年建置史和1400多年建城史，其发展始于战国时期，历经陆上贸易重镇的辉煌，至清朝金口港兴盛，逐渐蜕变为海陆交通枢纽。改革开放推动之下，即墨作为青岛下属县市，接受青岛辐射，商贸物流业和民营经济蓬勃发展，成为北方重要商贸名城。2017年即墨撤县设区，遵循习近平新时代中国特色社会主义思想，融入青岛都市区和胶东经济圈发展，抢抓国家战略机遇，规划2035年发展新路径，构建胶东经济圈一体化协同发展框架，构建高质量发展新空间格局，为第二个百年目标打下坚实基础。国土空间规划编制为党中央、国务院重要部署，旨在实现“多规合一”。即墨区国土空间分区规划在青岛市国土空间总体规划指导下，完成空间发展指南，勾勒即墨未来空间蓝图，成为开发、保护和建设活动基本依据。本手册以简洁图文，描绘即墨国土空间发展蓝图，让您了解即墨未来建设目标和发展方向。欢迎莅临即墨，共同见证2035年美好愿景的实现。\end{CJK*}     \\ \hline
ChatGPT  &   \begin{CJK*}{UTF8}{gbsn}
即墨市是一个有着千年历史的商都，拥有7000多年的人类活动史、2300多年的建置史和1400多年的建城史。该城市的发展始于战国时期，是陆上贸易重镇，后来随着清朝金口港的兴盛，逐渐成为海陆交通枢纽。随着改革开放的推进，即墨作为青岛的下辖县市，接受了青岛的辐射，商贸物流产业迅速发展，民营经济也随之崛起，逐渐成为中国北方的重要商贸名城。2017年即墨撤市设区后，我们遵循习近平新时代中国特色社会主义思想，全面推动融入青岛都市区和胶东经济圈的发展，抓住国家重大战略决策机遇，全面规划到2035年的发展新路径，积极融入胶东经济圈一体化协同发展框架，构建高质量发展的新空间格局，为实现第二个百年目标奠定坚实基础。国土空间规划编制是党中央、国务院的重要部署，旨在将各种空间规划融合在一起，实现“多规合一”。即墨区国土空间分区规划是在青岛市国土空间总体规划指导下完成的空间发展指南，是即墨未来发展的空间蓝图，也是各种开发、保护和建设活动的基本依据。本手册力求以简洁的图文语言，描绘即墨国土空间的发展蓝图，让您更好地了解即墨未来的建设目标和发展方向。来即墨，共同见证2035的美好。 \end{CJK*}     \\ \hline
\end{tabularx}}
\caption{Planning style transfer comparison }
\label{format} 
\end{table*}

\begin{table}[h]
\centering

\label{tab:data}
\caption{Data Classification and Description}
\begin{tabular}{|p{2.5cm}|p{5cm}|p{2.5cm}|p{4cm}|}
\hline
\textbf{Data Category} & \textbf{Data Description} & \textbf{Data Volume} & \textbf{Remarks} \\
\hline
Provincial Land Spatial Planning & Overall layout and guidance for a specific province, including strategies for the allocation, utilization, and management of various resources such as land, water, minerals, and forests. & Includes 29 provincial land spatial planning texts & Shanghai and Beijing have the latest urban master plans \\
\hline
Municipal Land Spatial Planning & Comprehensive planning for specific cities or municipal administrative regions, providing detailed guidance on the location, area, and use of various types of land. & Includes 337 municipal-level documents & Hong Kong has plans such as Hong Kong 2030+ and Northern Metropolis Area Plan \\
\hline
National Land Spatial Master Plan & Comprehensive planning at the national level, based on the country's development strategy and goals, coordinating and managing the national land spatial freedom. & 2820 planning-related case studies & Macau has the Macau 2040 Urban Master Plan \\
\hline
Spatial Planning Manuals & Includes research reports, policy recommendations, and planning proposals related to overall land spatial layout, regional coordinated development, providing decision-making basis for relevant departments. & Over 3000 planning texts at various administrative levels, case studies, and related Q\&A & Open source on the internet and compiled by various planning organizations. Planning Cloud website. \\
\hline
Authoritative Textbooks in the Field of Planning & Approximately 200 textbooks covering urban planning, remote sensing control, regional management, and traffic engineering for undergraduate and graduate students. These textbooks encompass the complete education of urban and rural planning at the postgraduate level. & Total of 1GB of text data in PDF version & Source: Baidu Wenku, GitHub, Teaching Syllabus \\
\hline
Some District and County-level Land Spatial Master Plans & Land spatial planning for district and county-level administrative areas, involving resource allocation, infrastructure planning, and past versions of planning documents drafted by relevant government departments at various levels, providing guidance and strategies for local development. & Supplementary documents for county-level planning texts & Source: Spatial Planning Manuals website \\
\hline
Past Provincial, County, and City Land Spatial Planning Texts (2000, 2010) & Including land spatial planning texts for provinces, counties, and cities in the years 2000 and 2010. & Total of 30GB of historical planning text data & Source: Compiled from Zhihu, including municipal, county, and village-level literature \\
\hline

\end{tabular}

\end{table}

\label{data}

\end{document}